\documentclass[letterpaper]{article} % DO NOT CHANGE THIS
\PassOptionsToPackage{table,dvipsnames}{xcolor}
\usepackage{aaai2026}  % DO NOT CHANGE THIS
\usepackage{times}  % DO NOT CHANGE THIS
\usepackage{helvet}  % DO NOT CHANGE THIS
\usepackage{courier}  % DO NOT CHANGE THIS
\usepackage[hyphens]{url}  % DO NOT CHANGE THIS
\usepackage{graphicx} % DO NOT CHANGE THIS
\usepackage{natbib}  % DO NOT CHANGE THIS AND DO NOT ADD ANY OPTIONS TO IT
\usepackage{caption} % DO NOT CHANGE THIS AND DO NOT ADD ANY OPTIONS TO IT
\usepackage{amsmath, amssymb, amsthm} 

\usepackage{newtxtext}
\usepackage{newtxmath}
\nocopyright

\usepackage{xcolor}
\usepackage{booktabs}
\usepackage{multirow}
\usepackage{pifont}
\usepackage{subcaption}
\usepackage{listings}
\usepackage[skins,listings]{tcolorbox}

\newcommand{\pmstd}[1]{{\scriptsize$\pm #1$}}

\newcommand{\xmark}{\ding{55}\ }

\theoremstyle{remark}
\newtheorem{definition}{Definition}

\providecommand{\pmstd}[1]{{\scriptsize$\pm #1$}}
\providecommand{\pmstdb}[1]{{\scriptsize\boldmath$\pm #1$}}

\definecolor{pw1}{HTML}{D7312D}
\definecolor{pw2}{HTML}{F2724D}
\definecolor{pw3}{HTML}{FEE395}
\definecolor{pw4}{HTML}{FEF9B7}
\definecolor{pw5}{HTML}{ACD2E5}
\definecolor{pw6}{HTML}{6090C1}
\definecolor{pw7}{HTML}{2B7DFC}
\definecolor{pw8}{HTML}{1B8EFF}
\definecolor{ranksecond}{HTML}{F3D7DA}
\definecolor{rankthird}{HTML}{DCE9F8}
\newcommand{\secondrank}[1]{\cellcolor{ranksecond}#1}
\newcommand{\thirdrank}[1]{\cellcolor{rankthird}#1}

\newcommand{\ProWorld}{%
  \textcolor{pw1}{P}%
  \textcolor{pw2}{r}%
  \textcolor{pw3}{o}%
  \textcolor{pw4}{W}%
  \textcolor{pw5}{o}%
  \textcolor{pw6}{r}%
  \textcolor{pw7}{l}%
  \textcolor{pw8}{d}%
}

\definecolor{codebg}{HTML}{F7F7FA}
\definecolor{codeblue}{HTML}{6A9ECF}
\definecolor{codered}{HTML}{E74C3C}
\definecolor{codegreen}{HTML}{76B77A}
\definecolor{codegray}{HTML}{8C8F96}
\lstdefinelanguage{GeoPython}{
  morekeywords={def,return},
  sensitive=true,
  morecomment=[l]{\#},
  morestring=[s]{"""}{"""}
}
\newtcblisting{GeoCodeBlock}{
  enhanced,
  colback=codebg,
  colframe=codebg,
  boxrule=0pt,
  arc=8pt,
  outer arc=8pt,
  boxsep=0pt,
  left=6pt,
  right=6pt,
  top=5pt,
  bottom=5pt,
  width=\linewidth,
  listing only,
  listing options={
    language=GeoPython,
    basicstyle=\fontsize{7.7}{8.8}\selectfont\ttfamily,
    keywordstyle=\color{codeblue},
    stringstyle=\color{codered},
    commentstyle=\color{codegreen}\itshape,
    identifierstyle=\color{codegray},
    emph={GeoWorldStyle,load_official_lewm_checkpoint,encoder,predictor,mse,mean,lorentz_project,hyperbolic_energy,triangle_reg,ProWorldDynamics,ProgressEntailment,sample_hindsight_ordered_pair,ProgressAwarePlanning,hyperbolic_predict,hyperbolic_dist,contrastive_future_loss,adaptive_cone,radius,cone_angle,CEM,rollout_cost},
    emphstyle=\color{codeblue},
    numbers=none,
    frame=none,
    xleftmargin=0pt,
    aboveskip=0pt,
    belowskip=0pt,
    columns=fullflexible,
    keepspaces=true,
    showstringspaces=false,
    breaklines=true,
    breakatwhitespace=true,
    tabsize=4
  }
}

\title{
\ProWorld: Progress-Aware Hyperbolic World Models for Long-Horizon Visual Goal Reaching
}
\author {
    Zihan Liu\textsuperscript{1,*},
    Yuzhe Zhuang\textsuperscript{1,*},
    Yuanzu Li\textsuperscript{1},
    Wanshuang Gou\textsuperscript{4},
    Jiahong Liu\textsuperscript{2},
    Min Zhou\textsuperscript{3},
    Menglin Yang\textsuperscript{1,\textdagger}
}
\affiliations {
    \textsuperscript{1} The Hong Kong University of Science and Technology (Guangzhou) \\
    \textsuperscript{2} The Chinese University of Hong Kong \\
    \textsuperscript{3} Yinwang Technologies Ltd.\\
    \textsuperscript{4} Independent Researcher \\
    \textsuperscript{*} Equal contribution. \textsuperscript{\textdagger} Corresponding author.
}

\begin{document}

\maketitle

\begin{abstract}
JEPA-style visual world models offer an effective paradigm for visual goal planning by predicting future latent representations. 
Existing methods typically learn local transition consistency through next-step representation prediction.
However, in long-horizon tasks,
accurate local prediction alone need not ensure sustained progress toward the goal.  
\textbf{First}, multi-step rollouts can remain locally plausible while drifting away from goal-relevant trajectories.
% existing methods still face two key challenges in long-horizon visual planning tasks. 
% First, alignment of consecutive-step representations provides no guarantee that multi-step rollouts consistently advance toward the goal. 
% Second, locally similar future states may be conflated in latent space despite carrying substantially different implications for long-term goal progress.
\textbf{Second}, locally similar future states can correspond to substantially different long-term progress, making them difficult to distinguish in a latent space optimized mainly for local consistency.
To address these challenges, we introduce goal-conditioned progress order, a relative ordering of states according to how they advance toward a given goal.
% We refer to the degree to which a state advances toward a given goal
% as its goal-conditioned progress order. 
This order exhibits an asymmetric, coarse-to-fine structure: early states retain broader future possibilities, while later states concentrate on more specific goal-relevant regions. 
Such a structure is well suited to hyperbolic geometry.
Motivated by this observation, we propose \textbf{ProWorld}, a progress-aware hyperbolic visual world model. 
ProWorld leverages goal-conditioned progress order to organize visual latent-space dynamics, maintains directional progress within trajectories via hyperbolic entailment learning, and mitigates progress ambiguity among locally similar future states via hyperbolic future discrimination. 
Furthermore, we design a progress-aware planning objective that scores candidate rollouts by jointly considering proximity to the goal and sustained progress across intermediate states. Experiments on four visual goal-reaching tasks demonstrate that ProWorld achieves an average absolute success-rate gain of 9.67 over LeWM. The code will be released after the paper is accepted.
\end{abstract}

\section{Introduction}
\label{sec:introduction}
\begin{figure}[t]
\centering
\includegraphics[width=\linewidth]{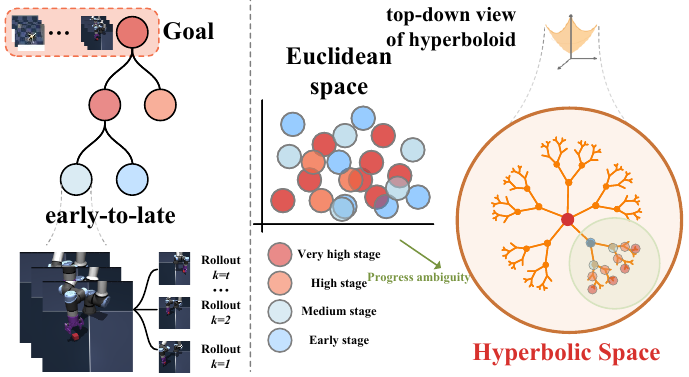}
\caption{Core motivation of ProWorld. In Euclidean latent space, locally similar future states may be mapped to nearby regions, causing progress ambiguity in long-horizon goal-directed planning. ProWorld organizes goal-conditioned progress in a hyperbolic latent space, where low-progress early states correspond to wider future possibilities while high-progress later states progressively contract toward more specific goal-relevant regions.}
\label{fig:intro_overview}
\end{figure}
Visual goal planning tasks require an agent to derive sequential action policies from raw pixel-level sensory inputs so that its state evolves toward a given goal through a series of intermediate states~\cite{finn2017deep,watter2015embed,hafner2019planet}. Recently, JEPA-style world models have established a principled, computationally efficient paradigm to address this task by forecasting future latent-space representations~\cite{ha2018worldmodels,lecun2022path,assran2023ijepa}. 
Typical approaches following this paradigm, such as LeWM~\cite{lewm} acquire local state transition invariance via one-step-ahead latent prediction, then leverage the inferred latent dynamics to conduct planning. Such approaches have yielded compelling empirical performance across short-horizon visual planning benchmarks~\cite{balestriero2025lejepa,lewm}.

% Visual goal planning requires an agent to select a sequence of actions based on pixel observations so that its state evolves to- ward a given goal through a series of intermediate states

%Representative methods such as LeWorldModel (Maes et al. 2026) typically learn local transi-tion consistency through next-step representation prediction and use the learned latent dynamics for planning, demon- strating promising effectiveness in short-horizon planning tasks

% learn latent-space dynamics end-to-end from pixels, constructing a visual world model suitable for planning via next-step representation prediction and SIGReg regularization, and have demonstrated their effectiveness on 
% While effective in short-horizon planning tasks~\cite{balestriero2025lejepa,lewm}.

However, strong performance on short-horizon planning tasks cannot guarantee robust planning capabilities for long horizons. In long-horizon settings,  goal states cannot be attained via a limited sequence of local actions. Instead, agents need to navigate through numerous intermediate states and sustain goal-aligned progress across extended time spans~\cite{pertsch2020longhorizon,gieselmann2023expansive,ogbench}. This exposes two core challenges. \textbf{First}, JEPA-style world models typically learn local transition consistency by aligning predicted future representations with ground truths. Nevertheless, this alignment loss only enforces fidelity for the very next immediate state. A potential objective mismatch between one-step prediction quality and downstream control performance~\cite{lambert2020objective,nair2020goalaware} makes it difficult to guarantee that multi-step rollouts obtained by recursive unrolling exhibit sustained goal-directed progress. Consequently, the model may generate intermediate states that are locally plausible in terms of dynamics yet globally ineffective with respect to the goal, a problem that compounds with rollout horizon due to accumulating model bias~\cite{janner2019mbpo,pertsch2020longhorizon}. 
\textbf{Second}, future states arising in different trajectory contexts may appear similar in visual appearance or local transition structure yet carry different goal-progress semantics; if such states are mapped too closely in latent space, the planner may confuse locally plausible but globally divergent candidate futures and select action sequences that are globally ineffective~\cite{eysenbach2022contrastive}.

% In long-horizon planning, therefore, learning only locally predictable latent dynamics is insufficient. The model must also capture, given a goal, which states represent an earlier exploratory phase and which represent a phase of advancing closer to the goal. 
Therefore, long-horizon visual planning requires more than locally predictable latent dynamics. Given a goal, a model is expected to also learn to sort states according to their relative progress toward achieving that goal.
We  define this structural property as \textbf{goal-conditioned progress order}, a relation that describes how states in a goal-conditioned hindsight trajectory segment evolve from lower-progress stages to higher-progress stages.
%This relation is asymmetric and coarse-grained: it does not require every step to strictly reduce the distance to the goal, but instead captures a progress trend jointly shaped by temporal order, goal proximity, and reachability. 
This relation possesses two key attributes: it is asymmetric and coarse-grained. Rather than mandating strict goal-distance reduction at every individual timestep, it encapsulates the overall progress trend collectively determined by temporal sequence, proximity to the target goal, and state reachability. Such asymmetric, coarse-to-fine hierarchical structure aligns naturally with hyperbolic geometry, which excels at compactly embedding hierarchical relational patterns.
% Specifically, within a goal-conditioned hindsight trajectory segment, states of lower progress should gradually evolve into states of higher progress, forming an asymmetric progress relation. 
% This relation does not require every step in the trajectory to strictly reduce the distance to the goal; rather, it characterizes a coarse-grained progress trend jointly induced by temporal order, goal proximity, and reachability.

\textbf{Proposed Method.}
Building on these observations, we propose ProWorld, a progress-aware hyperbolic world model for long-horizon visual planning, whose core motivation is illustrated in Fig.~\ref{fig:intro_overview}. In brief, ProWorld maps visual representations to the Lorentz model of hyperbolic space~\cite{nickel2018lorentz,ganea2018hyperbolic} and uses goal-conditioned progress order as a geometric inductive bias for organizing latent-space dynamics. 
\textbf{During training}, ProWorld employs hyperbolic entailment learning, which encourages states within the same trajectory to unfold along goal-directed progress directions while hyperbolic future discrimination separates locally similar future states with different progress semantics.
% to constrain states within the same trajectory to unfold along goal-directed progress directions, thereby alleviating the lack of sustained progress in multi-step rollouts; 
% simultaneously, hyperbolic future discrimination enhances the separability of future states arising in different trajectory contexts, reducing confusion in latent space between candidate futures that are locally similar but carry different progress semantics. 
\textbf{During planning}, ProWorld evaluates candidate action sequences using a progress-aware trajectory cost that considers both goal proximity and progress consistency across intermediate states.
% adopts a trajectory cost consistent with this progress structure, so that candidate action sequences are evaluated not only by their terminal distance to the goal but also by whether intermediate states consistently advance in an effective direction. 
We evaluate ProWorld across four visual goal-reaching tasks from the OGBench benchmark suite~\cite{ogbench}. 
Across all settings, ProWorld attains an average absolute success rate (SR) improvement of 9.67 compared with LeWM.

The contributions of this paper are summarized as follows:
\begin{itemize}
    \item We introduce goal-conditioned progress order as a structural view of long-horizon visual goal planning, and instantiate it in ProWorld by organizing visual latent-space dynamics in the Lorentz model of hyperbolic space. This provides a geometric inductive bias for modeling the asymmetric coarse-to-fine relation by which states evolve from lower-progress stages to higher-progress, goal-relevant regions.

    % \item We design a hyperbolic entailment learning constraint that uses adaptive entailment cones to jointly model intra-trajectory directional consistency and radial hierarchical structure, thereby encouraging latent states to evolve along goal-relevant progress directions.
    \item We introduce a progress-aware learning framework that jointly models directional progress within trajectories and discriminability across future contexts, addressing both rollout drift and progress ambiguity in long-horizon planning.

    % \item We develop two progress-aware learning objectives: hyperbolic entailment learning for maintaining intra-trajectory directional progress, and hyperbolic future discrimination for separating locally similar futures with different long-term progress semantics.

    \item Experiments on four visual goal-planning tasks show that ProWorld consistently improves long-horizon planning performance, achieving an average absolute SR gain of 9.67 over LeWM. 
    % We open-source the code, training configurations, and evaluation protocols to provide a reproducible foundation for future research on visual goal planning.
\end{itemize}

\section{Related Work}
\subsection{Visual World Models for Goal-Conditioned Planning}
\textbf{Visual world models.} Visual world models learn predictive environment dynamics for control and planning. Pixel-space prediction~\cite{finn2017deep} was subsequently complemented by compact latent dynamics, as in PlaNet~\cite{hafner2019planet} and Dreamer~\cite{hafner2020dreamer}. Task-oriented latent models such as MuZero~\cite{schrittwieser2020muzero} and TD-MPC~\cite{hansen2022tdmpc} further show that decision-relevant dynamics need not reconstruct pixels accurately. Nevertheless, recursive rollouts remain vulnerable to accumulated model bias~\cite{janner2019mbpo} and mismatch between predictive objectives and downstream control~\cite{lambert2020objective}.

\textbf{JEPA-style prediction.} JEPA-style models shift the focus from pixel reconstruction to latent prediction. I-JEPA~\cite{assran2023ijepa} and V-JEPA~\cite{bardes2024vjepa} establish this paradigm for images and videos, while LeJEPA~\cite{balestriero2025lejepa} introduces SIGReg for stable representation learning. LeWM~\cite{lewm} extends this formulation to action-conditioned visual dynamics. DINO-WM~\cite{Dino-wm} and Causal-JEPA~\cite{c-jepa} further investigate pretrained features and object-centric prediction for planning. These methods primarily emphasize predictive accuracy or local transition structure, whereas ProWorld explicitly organizes latent dynamics by goal-conditioned progress.

\textbf{Long-horizon planning.} Long-horizon goal reaching requires more than locally accurate prediction. SoRB~\cite{eysenbach2019sorb} performs graph-based search over replay states, hierarchical visual planning~\cite{pertsch2020longhorizon} predicts intermediate subgoals, and contrastive goal-conditioned learning~\cite{eysenbach2022contrastive} models state-goal controllability. Unlike policy learning or explicit high-level search, ProWorld embeds progress structure directly within an action-conditioned world model and uses it to evaluate recursively predicted trajectories.

% \subsection{Hyperbolic and Geometric Representation Learning}
% Hyperbolic spaces are well suited for representing coarse-to-fine structure, where broad abstract regions progressively branch into more specific regions.
% Hyperbolic spaces provide low-distortion representations for hierarchical data, as demonstrated by Poincar\'e embeddings~\cite{nickel2017poincare}. The Lorentz model~\cite{nickel2018lorentz} offers favorable optimization properties, while Hyperbolic Entailment Cones~\cite{ganea2018hyperbolic} encode asymmetric partial orders through directional cones. These properties make hyperbolic geometry a natural carrier for the coarse-to-fine and asymmetric progress relations studied in this paper.

% GeoWorld~\cite{zhang2026geoworld} also introduces hyperbolic geometry into JEPA-style world models, combining V-JEPA 2 representations with Geometric Reinforcement Learning to reduce geometric degradation during long rollouts. ProWorld instead models goal-conditioned progress order and uses the resulting geometry in both representation learning and progress-aware planning.

\subsection{Hyperbolic and Geometric Representation Learning}
\textbf{Coarse-to-fine geometry.} Hyperbolic spaces provide a natural geometry for coarse-to-fine organization, where broad abstract regions progressively branch into more specific regions with low distortion. This property has been widely explored in visual and multimodal representation learning, including image embeddings~\cite{khrulkov2020hyperbolic}, zero-shot recognition~\cite{liu2020hyperbolicvisual}, segmentation~\cite{atigh2022hyperbolicseg}, metric learning~\cite{ermolov2022hyperbolic}, image-text representation learning~\cite{desai2023meru}, and scene-object representation learning~\cite{ge2023hyperboliccontrastive}. Recent work further extends this perspective to process-structured settings, such as representation evolution over time~\cite{bui2025arrow}, dense retrieval with general-to-specific semantic structure~\cite{madhu2026hyprag}, sequential modeling~\cite{sohn2022bending,zhang2025hmamba}, deep reinforcement learning~\cite{cetin2022hyperbolicrl,klein2026hyperplusplus}, and multi-step reasoning~\cite{liu2026hyperguide}. These works suggest that hyperbolic geometry is useful when representations must capture both hierarchical specificity and branching future possibilities.

\textbf{Hyperbolic world models.}
GeoWorld~\cite{zhang2026geoworld} also introduces hyperbolic geometry into JEPA-style world models to reduce geometric degradation during long rollouts. ProWorld shares the motivation of using hyperbolic geometry for long-horizon world modeling, but explicitly models goal-conditioned progress order, organizing states from broad, lower-progress regions toward more specific goal-relevant regions. This progress structure is used not only for representation learning through entailment and future discrimination, but also for progress-aware planning over candidate action sequences.

\section{Preliminaries}
\label{sec:preliminaries}

\subsection{The Lorentz Model of Hyperbolic Space}
\label{sec:prelim_lorentz}

To model the hierarchical structure of long-horizon trajectories, ProWorld embeds state representations into the Lorentz model of hyperbolic space with negative curvature. We introduce the basic geometric notation used throughout this paper. Let $m$ denote the dimensionality of the hyperbolic latent space and $c>0$ denote the curvature parameter, so that the corresponding hyperbolic space has negative curvature $-c$. For any point $x \in \mathbb{R}^{m+1}$, let $x_0$ denote its time component and $x_{1:m}$ its spatial components. The $m$-dimensional hyperbolic space in the Lorentz model is defined as
\begin{equation}
    \mathbb{H}^{m}_{c}
    =
    \left\{
    x \in \mathbb{R}^{m+1}
    \quad\middle|\quad
    \langle x,x\rangle_{\mathcal{L}} = -\frac{1}{c}
    \quad
    x_0 > 0
    \right\}
\end{equation}
where $\langle \cdot,\cdot\rangle_{\mathcal{L}}$ denotes the Lorentz inner product. For any $x,y \in \mathbb{R}^{m+1}$, this inner product is defined as
\begin{equation}
    \langle x,y\rangle_{\mathcal{L}}
    =
    \sum_{i=1}^{m} x_i y_i - x_0 y_0
\end{equation}
Given two hyperbolic points $x,y \in \mathbb{H}^{m}_{c}$, the geodesic distance between them is
\begin{equation}
    d_{\mathbb{H}}(x,y)
    =
    \frac{1}{\sqrt{c}}
    \operatorname{arcosh}
    \left(
    -c \langle x,y\rangle_{\mathcal{L}}
    \right)
\end{equation}
where $d_{\mathbb{H}}(\cdot,\cdot)$ denotes the hyperbolic geodesic distance and $\operatorname{arcosh}(\cdot)$ denotes the inverse hyperbolic cosine function.

\subsection{Goal-Conditioned Progress Order}
\label{sec:goal_conditioned_progress}

Long-horizon visual planning requires the model to capture not only local transition dynamics but also how states progress toward a given goal. In successful goal-conditioned trajectories, earlier states usually correspond to lower progress and broader future possibilities, whereas later states correspond to higher progress and more goal-specific regions. We formalize this structure as follows.

\begin{definition}[Goal-conditioned progress order]
\label{def:progress_order}
Consider a goal-conditioned hindsight trajectory segment
$\tau=(s_0,s_1,\ldots,s_T;g)$ with states $s_t\in\mathcal{S}$ and goal $g\in\mathcal{G}$. Let $\rho_g:\mathcal{S}\rightarrow\mathbb{R}$ denote an underlying progress potential, where larger $\rho_g(s)$ indicates higher progress of state $s$ toward goal $g$. For two states $s_i$ and $s_j$ in the same trajectory, we define
\[
s_i \prec_g s_j
\quad \text{if} \quad
i<j
\quad \text{and} \quad
\rho_g(s_j)\geq \rho_g(s_i)+\epsilon
\]
where $\epsilon>0$ is a progress margin.
\end{definition}

Since $\rho_g$ is generally unobserved, we use temporally separated state pairs from goal-conditioned hindsight trajectories as weak supervision: if $j-i\geq\Delta_{\min}$, we treat $(s_i,s_j;g)$ as an empirical relation $s_i\prec_g s_j$. This construction does not require every adjacent transition to strictly reduce goal distance; instead, it applies a soft progress constraint to temporally separated pairs that are expected to move toward the hindsight goal. \textbf{Additional discussion on asymmetry and cone realization is provided in the Appendix.}

\begin{figure}[t]
\centering
\includegraphics[width=1.0\linewidth]{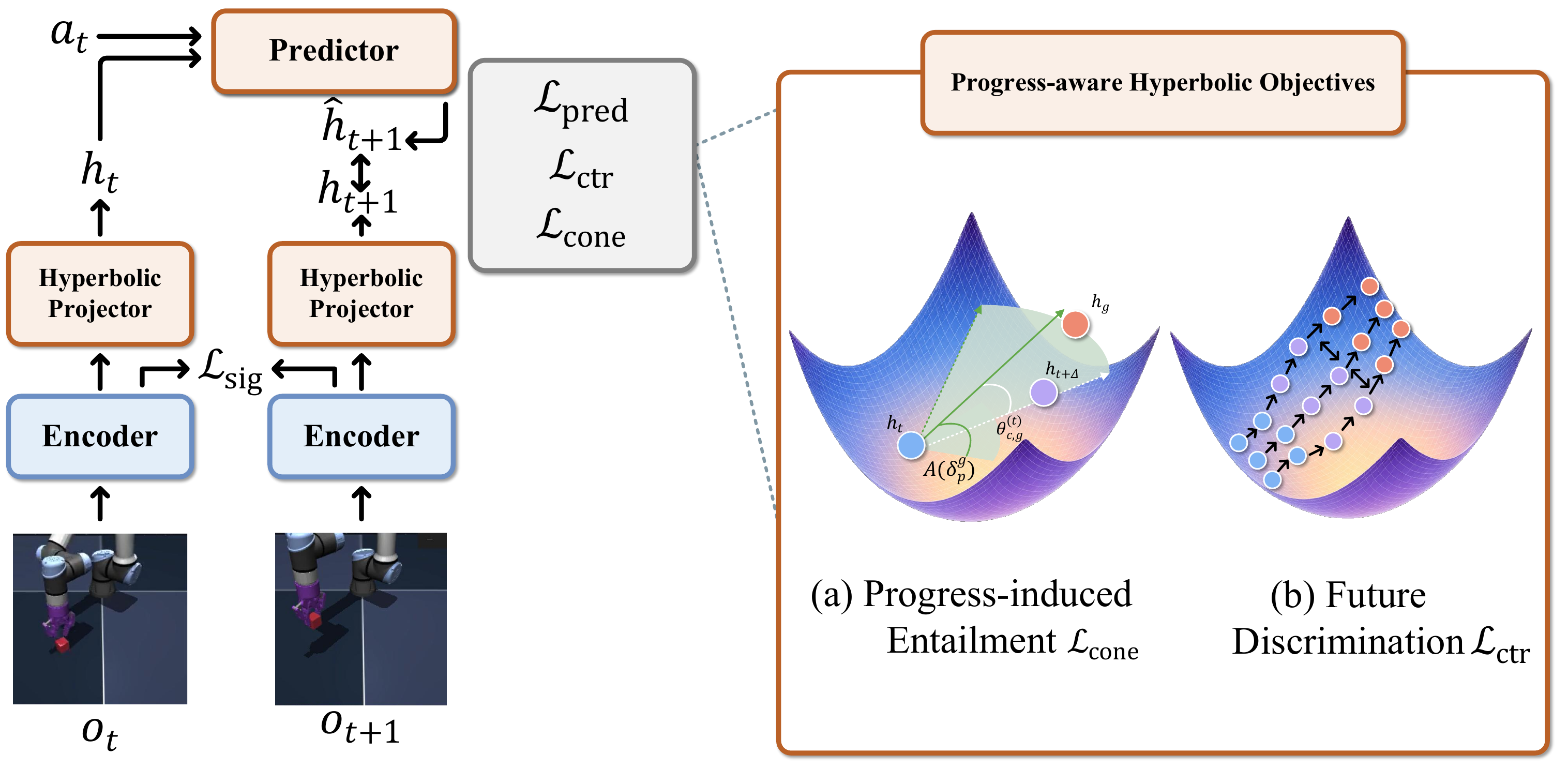}
\centerline{(a) Training stage}
\includegraphics[width=1.0\linewidth]{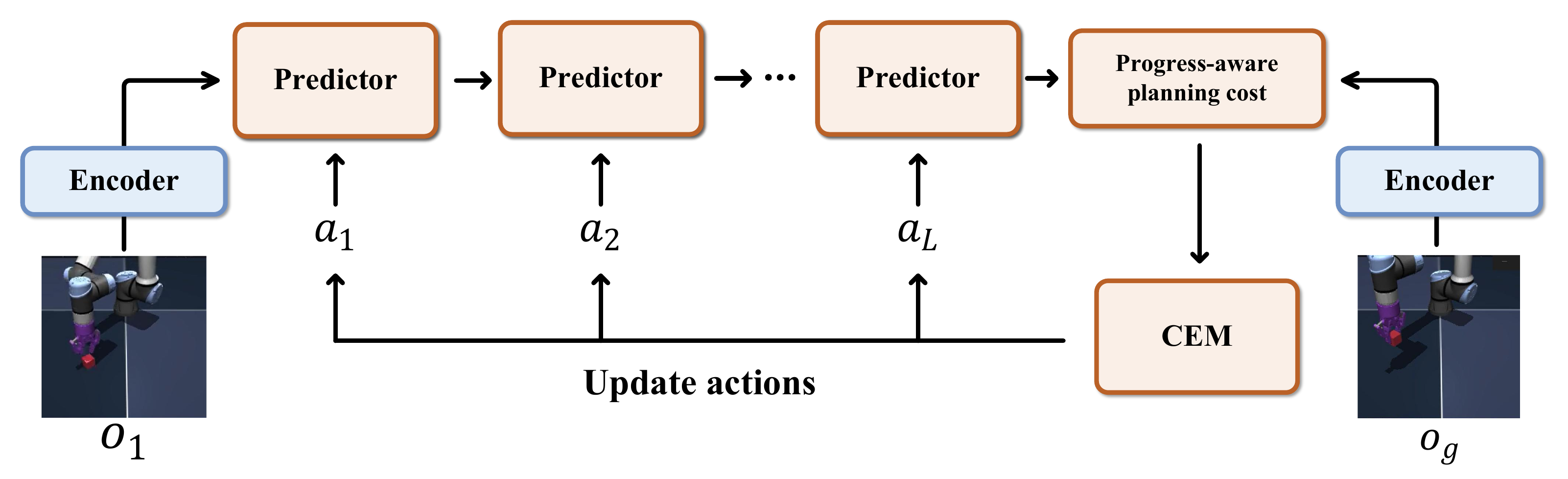}
\centerline{(b) Planning stage}
\caption{
Overview of ProWorld, including progress-aware hyperbolic dynamics learning and CEM-based latent planning.
}
\label{fig:method_overview}
\end{figure}

\section{Method}
\label{sec:method}

\subsection{Overview}
\label{sec:method_overview}

Given offline trajectories $\tau=\{(o_t,a_t)\}_{t=1}^{T}$, ProWorld learns a latent dynamics model for goal-conditioned visual planning. As shown in Fig.~\ref{fig:method_overview}, pixel observations are encoded into Euclidean latents, mapped to the Lorentz hyperbolic space, and predicted forward by an action-conditioned autoregressive predictor. The hyperbolic space is used to impose progress-aware geometric supervision during training and to score imagined rollouts during planning.

Training converts temporal relations and encoded goals in goal-conditioned hindsight trajectories into progress-aware constraints, encouraging the latent dynamics to capture both local predictability and directional progress. At inference time, the learned model rolls out candidate action sequences and guides CEM using a progress-aware hyperbolic planning cost.

\subsubsection{Visual Encoding and Hyperbolic Embedding}
\label{sec:hyperbolic_latent}
Given a pixel observation $o_t$, the visual encoder $f_\theta$ and Euclidean projection head $g_\psi$ first produce
\begin{equation}
    z_t = g_\psi(f_\theta(o_t)) \in \mathbb{R}^{d}
\end{equation}
The hyperbolic projection head $q_\omega$ maps this Euclidean latent to an $m$-dimensional spatial tangent component at the Lorentz origin,
\begin{equation}
    \tilde{u}_t = q_\omega(z_t) \in \mathbb{R}^{m}
\end{equation}
Following this tangent-space construction, we normalize by feature dimension and smoothly clip the tangent norm:
\begin{equation}
\label{eq:6}
    \bar{u}_t = \tilde{u}_t / \sqrt{m}
    \qquad
    u_t =
    \frac{\rho_{\max}\tanh\left(\|\bar{u}_t\|_2 / \rho_{\max}\right)}
    {\|\bar{u}_t\|_2 + \varepsilon_{\mathrm{num}}}
    \bar{u}_t
\end{equation}
Here $\rho_{\max}$ is the tangent-norm bound, set to $4.0$ by default, and $\varepsilon_{\mathrm{num}}$ is a small constant for numerical stability.
The stabilized spatial component $u_t$ is then lifted to the Lorentz tangent space by appending a zero time component, yielding $(0,u_t)\in\mathbb{R}^{m+1}$. The hyperbolic latent representation is obtained by the origin exponential map:
\begin{equation}
    h_t =
    \exp^{c}_{\mathbf{o}_c}\left((0,u_t)\right)
    \in \mathbb{H}_{c}^{m}
    \subset \mathbb{R}^{m+1}
\end{equation}
\paragraph{Action-conditioned autoregressive predictor.}
Given an action $a_t$, the action encoder produces $\alpha_t=e_\eta(a_t)$. With a history window of length $H$, the autoregressive predictor forecasts the next Euclidean latent as
\begin{equation}
    \hat{z}_{t+1}
    =
    p_\phi(z_{t-H+1:t}, \alpha_{t-H+1:t})
\end{equation}
where $\phi$ denotes the predictor parameters and $\hat{z}_{t+1}$ is the predicted next latent. The predictor is implemented with a Transformer and conditioned on action embeddings. The prediction is mapped to the Lorentz manifold using the same dimension normalization, smooth clipping, and zero-time tangent lifting as in Eq.~\ref{eq:6}. Specifically, we first compute $\hat{\tilde{u}}_{t+1}=q_\omega(\hat{z}_{t+1})$ and obtain $\hat{u}_{t+1}$ by applying the transformation in Eq.~(6), then
\begin{equation}
    \hat{h}_{t+1}
    =
    \exp^c_{\mathbf{o}_c}\left((0,\hat{u}_{t+1})\right)
\end{equation}
where $\hat{h}_{t+1}$ is the predicted future hyperbolic point.

\subsection{Progress-Aware Hyperbolic Dynamics Learning}
\label{sec:training_objective}

\subsubsection{Hyperbolic Prediction and Discrimination}
\label{sec:hyperbolic_pred_ctr}

ProWorld first aligns the predicted future state with the encoded true future state in the Lorentz model. Let $\hat{h}_{t+1}$ denote the hyperbolic point predicted from the current state and action, and let $h_{t+1}$ denote the hyperbolic point encoded from the true future observation. The prediction loss is
\begin{equation}
    \mathcal{L}_{\mathrm{pred}}
    =
    \mathbb{E}_{t}
    \left[
    d_{\mathbb{H}}(\hat{h}_{t+1}, h_{t+1})^2
    \right]
\end{equation}
where $\mathbb{E}_{t}$ denotes the empirical expectation over sampled training timesteps.

\begin{table*}[!h]
\centering
\caption{Comparison of success rates across different methods on the visual goal-conditioned planning task. All success rate values are reported on a 0-100 scale with the percent sign omitted. Results with $\pm$ indicate mean and standard deviation over training seeds 42, 1337, and 3407. Best results are bolded; second-best and third-best cells are shaded in light red and light blue, respectively. The last row reports the relative SR improvement over the second-best baseline.}
\label{tab:main_results}
\small
\providecommand{\pmstd}[1]{{\scriptsize$\pm #1$}}
\begin{tabular}{lcccc}
\toprule
\textbf{Method} & \textbf{PushT} & \textbf{Cube-S} & \textbf{AntMaze-L} & \textbf{Scene} \\
\midrule
LeWM~\cite{lewm}             & \secondrank{83.33\pmstd{4.04}} & \secondrank{71.33\pmstd{3.06}} & 22.00\pmstd{4.00} & 57.33\pmstd{2.49} \\
C-JEPA~\cite{c-jepa}         & 66.67\pmstd{3.77} & 48.00\pmstd{7.12} & 20.67\pmstd{0.94} & 48.00\pmstd{1.63} \\
GCIQL~\cite{ogbench} & 10.00\pmstd{3.27} & 62.00\pmstd{1.63} & \secondrank{23.33\pmstd{3.40}} & 40.67\pmstd{6.18} \\
GCIVL~\cite{ogbench}         & 14.00\pmstd{8.49} & \thirdrank{64.00\pmstd{4.32}} & 18.67\pmstd{2.49} & 45.33\pmstd{6.79} \\
PLDM~\cite{PLDM}             & \thirdrank{80.67\pmstd{2.49}} & 57.33\pmstd{2.49} & \thirdrank{22.67\pmstd{3.77}} & \thirdrank{58.67\pmstd{5.25}} \\
EB-JEPA~\cite{EB-JEPA}       & 77.33\pmstd{2.49} & 56.67\pmstd{1.89} & 19.33\pmstd{2.49} & \secondrank{67.33\pmstd{0.94}} \\
TD-MPC2~\cite{TD-MPC2}       & 4.00\pmstd{1.63} & 44.00\pmstd{4.32} & 18.67\pmstd{0.94} & 46.67\pmstd{3.40} \\
Sub-JEPA~\cite{Sub-JEPA}     & 77.33\pmstd{5.25} & 46.00\pmstd{0.00} & 19.33\pmstd{2.49} & \thirdrank{58.67\pmstd{4.71}} \\
GeoWorld-Style\textsuperscript{a}~\cite{zhang2026geoworld} & 72.67\pmstd{3.06} & 53.33\pmstd{4.16} & 18.00\pmstd{2.00} & 51.33\pmstd{2.49} \\
Random                       & 0.00 & 42.00 & 14.00 & 36.00 \\
\midrule
\rowcolor[HTML]{e6e6e6}
\textbf{ProWorld (Ours)}     & \textbf{94.00}\pmstdb{2.00} & \textbf{77.33}\pmstdb{1.15} & \textbf{32.00}\pmstdb{2.00} & \textbf{69.33}\pmstdb{0.94} \\
\textbf{Rel. Gain} & \textbf{+12.80\%} & \textbf{+8.41\%} & \textbf{+37.16\%} & \textbf{+2.97\%} \\
\bottomrule
\end{tabular}
\parbox{\linewidth}{\scriptsize \textsuperscript{a} GeoWorld-Style denotes our reproduction of GeoWorld under the unified experimental setting.}
\end{table*}

\begin{table*}[!h]
    \centering
    \begin{minipage}[t]{0.48\textwidth}
        \vspace{0pt}
        \centering
        \caption{Module ablation study of ProWorld on Cube-S. A blank entry indicates the same setting as the full model; \xmark indicates removal of the corresponding component. Best SR is bolded; second-best and third-best SR cells are shaded in light red and light blue, respectively.}
        \label{tab:ablation}
        \vspace{1mm}
        \small
        \resizebox{\linewidth}{!}{
        \begin{tabular}{lccccr}
            \toprule
            \textbf{Variant} & \textbf{Pred.} & \textbf{Ctr.} & \textbf{Cone.} & \textbf{Reg.} & \textbf{SR} \\
            \midrule
            \rowcolor[HTML]{e6e6e6} \textbf{Full}      &  &  &  &  & \textbf{78\%} \\
            w/o Cone.                   &  &  & \xmark &  & \thirdrank{62\%} \\
            w/o Ctr.                    &  & \xmark &  &  & 46\% \\
            w/o Reg.                    &  &  &  & \xmark & \secondrank{76\%} \\
            w/o Ctr./Cone          &  & \xmark & \xmark &  & 42\% \\
            w/o Ctr./Cone/Reg.   &  & \xmark & \xmark & \xmark & 46\% \\
            \bottomrule
        \end{tabular}
        }
    \end{minipage}
    \hfill
    \begin{minipage}[t]{0.48\textwidth}
        \vspace{0pt}
        \centering
        \caption{Ablation study on geometric structure and goal-conditioned progress order. A blank entry indicates the same setting as the full model; \xmark indicates removal or replacement of the corresponding design. Best SR is bolded; second-best and third-best SR cells are shaded in light red and light blue, respectively.}
        \label{tab:geometry_order_ablation}
        \vspace{1mm}
        \small
        \resizebox{\linewidth}{!}{
        \begin{tabular}{lccccr}
            \toprule
            \textbf{Variant} & \textbf{Hyp.} & \textbf{Correct} & \textbf{Same Traj.} & \textbf{Adaptive} & \textbf{SR} \\
             &  & \textbf{Order} & \textbf{Pair} & \textbf{Cone} & \\
            \midrule
            \rowcolor[HTML]{e6e6e6} \textbf{Full}      &  &  &  &  & \textbf{78\%} \\
            Euc.      & \xmark &  &  &  & \thirdrank{68\%} \\
            Rev.      &  & \xmark &  &  & 44\% \\
            Rand.     &  &  & \xmark &  & 52\% \\
            Fixed     &  &  &  & \xmark & \secondrank{70\%} \\
            \bottomrule
        \end{tabular}
        }
    \end{minipage}
\end{table*}

Pointwise prediction alone may entangle future states that are locally similar but have different progress semantics. To improve future-state discriminability, we introduce a Lorentz contrastive constraint. Given a batch of $N$ predicted points $\{\hat{h}_i\}_{i=1}^{N}$ and their matched true future points $\{h_i\}_{i=1}^{N}$, where $i$ indexes samples and $N$ is the batch size, we compute the pairwise Lorentz geodesic distance
\begin{equation}
    D_{ij}
    =
    d_{\mathbb{H}}(\hat{h}_i,h_j)
    =
    \frac{1}{\sqrt{c}}
    \operatorname{arcosh}
    \left(
    -c\langle \hat{h}_i,h_j\rangle_{\mathcal{L}}
    \right)
\end{equation}
where $D_{ij}$ is the distance between the $i$-th predicted point and the $j$-th true future point. For each $\hat{h}_i$, $h_i$ is the positive future sample and the remaining $h_j$ serve as negatives. The contrastive logits are
\begin{equation}
    \ell_{ij} = -\frac{D_{ij}}{\tau}
\end{equation}
where $\ell_{ij}$ is the $(i,j)$-th logit and $\tau>0$ is the temperature. We use a symmetric cross-entropy loss so that predictions retrieve their matched futures and true futures retrieve their matched predictions:
\begin{equation}
    \mathcal{L}_{\mathrm{ctr}}
    =
    \frac{1}{2}
    \left[
    \mathrm{CE}(\ell, y)
    +
    \mathrm{CE}(\ell^\top, y)
    \right]
\end{equation}
where $\ell^\top$ denotes the transposed logits for reverse matching, $\mathrm{CE}(\cdot,\cdot)$ denotes cross-entropy and $y_i=i$ denotes the within-batch matching label. This constraint pulls matched future pairs together while separating unmatched future states on the Lorentz manifold.

\subsubsection{Progress-Induced Entailment Constraint}
\label{sec:cone}

\begin{figure}[t]
\centering
\includegraphics[width=0.96\linewidth]{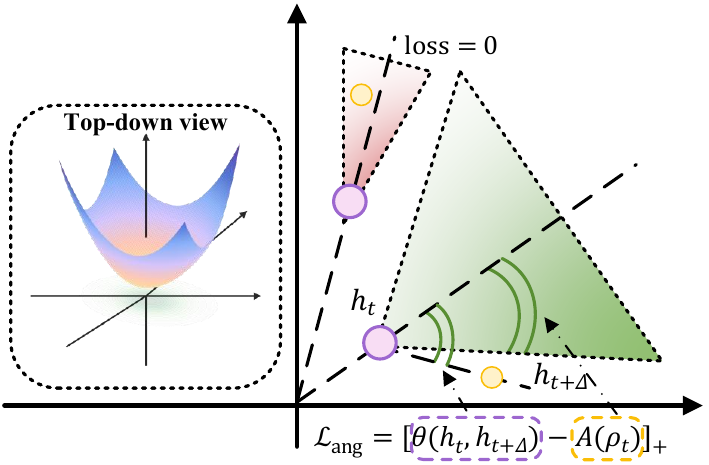}
\caption{
Illustration of the goal-anchored adaptive entailment cone. The green region denotes the zero angular-violation area, where the child direction from the parent remains aligned with the parent-to-goal direction.
}
\label{fig:cone}
\end{figure}

To realize goal-conditioned progress order in the hyperbolic latent space, ProWorld introduces a goal-anchored adaptive entailment cone constraint. Given a hindsight goal observation $o_g$, we encode it with the same hyperbolic projection pipeline to obtain $h_g$. For two states before this goal in the same trajectory with temporal offset $\Delta>0$, we treat the earlier state as the parent and the later state as the child:
\begin{equation}
    h^{p}_{t}=h_t
    \qquad
    h^{c}_{t}=h_{t+\Delta}
\end{equation}
Fig.~\ref{fig:cone} illustrates the resulting goal-anchored cone construction.

Instead of defining progress only with respect to the Lorentz origin, we compare two tangent directions at the parent point:
\begin{equation}
    v_c = \log^c_{h^p_t}(h^c_t)
    \qquad
    v_g = \log^c_{h^p_t}(h_g)
\end{equation}
where $v_c$ points from the parent to the child and $v_g$ points from the parent to the goal. After normalizing the tangent directions,
\begin{equation}
    \bar{v}_c = \frac{v_c}{\|v_c\|_{\mathcal{L}}}
    \qquad
    \bar{v}_g = \frac{v_g}{\|v_g\|_{\mathcal{L}}}
\end{equation}
we define the goal-conditioned angular deviation as
\begin{equation}
    \theta_{c,g}^{(t)}
    =
    \arccos
    \left(
    \langle \bar{v}_c,\bar{v}_g\rangle_{\mathcal{L}}
    \right)
\end{equation}
which measures whether the child moves along the direction from the current state toward the hindsight goal.

The cone aperture is adapted according to the normalized parent-to-goal distance
\begin{equation}
\begin{aligned}
    \delta_p^g =
    \operatorname{clip}
    \left(
    \frac{d_{\mathbb{H}}(h^p_t,h_g)}{\rho_{\max}},0,1
    \right)\\
    A(\delta_p^g)
    =
    A_{\min}
    +
    (A_{\max}-A_{\min})\delta_p^g
\end{aligned}
\end{equation}
where $A_{\min}$ and $A_{\max}$ are the minimum and maximum apertures. Farther states induce wider cones, while states closer to the goal impose finer directional constraints. The final entailment cone loss is
\begin{equation}
\begin{aligned}
    \mathcal{L}_{\mathrm{cone}}
    =
    \mathbb{E}_{t}
    \Big[
    &w_{\theta}
    \max(0, \theta_{c,g}^{(t)} - A(\delta_p^g)) \\
    &+
    w_{g}
    \max(0, d_{\mathbb{H}}(h^c_t,h_g)+m_g \\
    &\hspace{16mm}
    -d_{\mathbb{H}}(h^p_t,h_g))
    \Big]
\end{aligned}
\end{equation}
where $w_{\theta}$ and $w_g$ weight the angular and goal-distance terms, and $m_g$ is the goal-progress margin. The angular term keeps the child inside the parent-to-goal cone, while the distance term encourages the child to become closer to the hindsight goal than the parent. Thus, the same state can induce different progress directions under different goals through $h_g$.

\paragraph{Stabilization and overall objective.}
Following LeWM, we retain the SIGReg regularizer $\mathcal{L}_{\mathrm{sig}}$ to stabilize the latent distribution and prevent representation collapse. The final training objective is
\begin{equation}
\mathcal{L}
=
\lambda_{\mathrm{pred}}\mathcal{L}_{\mathrm{pred}}
+
\lambda_{\mathrm{ctr}}\mathcal{L}_{\mathrm{ctr}}
+
\lambda_{\mathrm{cone}}\mathcal{L}_{\mathrm{cone}}
+
\lambda_{\mathrm{sig}}\mathcal{L}_{\mathrm{sig}}
\end{equation}

\subsection{Hyperbolic Latent Planning}
\label{sec:planning}

During inference, the goal observation $o_g$ is encoded by the same pipeline to obtain the goal point $h_g\in\mathbb{H}^{m}_{c}$. For a candidate action sequence $\mathbf{a}_{1:L}=(a_1,\ldots,a_L)$ with rollout horizon $L$, ProWorld recursively predicts hyperbolic states $\{\hat{h}_t\}_{t=1}^{L}$ and measures their stepwise goal distances as
\begin{equation}
    D_t^g = d_{\mathbb{H}}(\hat{h}_t, h_g)^2
\end{equation}
where $D_t^g$ is the squared hyperbolic distance from the $t$-th predicted state to the goal. Besides the terminal distance $D_L^g$, we use the best and mean distances along the rollout:
\begin{equation}
\begin{aligned}
    C_{\mathrm{best}} &= \min_{1\leq t\leq L} D_t^g \\
    C_{\mathrm{mean}} &= \frac{1}{L}\sum_{t=1}^{L}D_t^g
\end{aligned}
\end{equation}
where $C_{\mathrm{best}}$ captures the closest intermediate progress and $C_{\mathrm{mean}}$ captures average trajectory proximity. The planning cost is
\begin{equation}
    C
    =
    \beta_T D_L^g
    +
    \beta_b C_{\mathrm{best}}
    +
    \beta_m C_{\mathrm{mean}}
\end{equation}
where $\beta_T$, $\beta_b$, and $\beta_m$ weight the terminal, best-intermediate, and mean-distance terms. ProWorld selects the action sequence minimizing $C$ and executes its first action.

\begin{figure}[t]
\centering
\includegraphics[width=\linewidth]{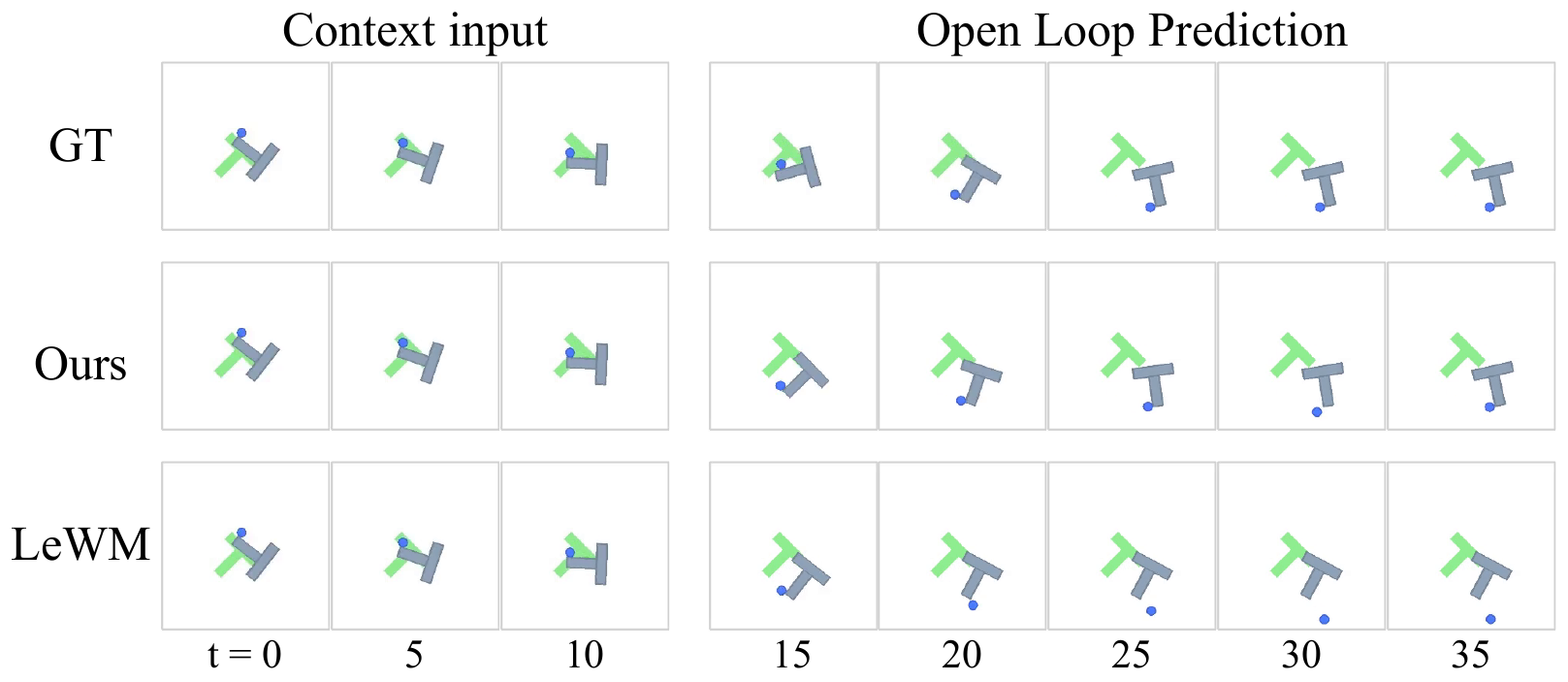}
\caption{Qualitative planning visualization on the PushT task. Rollouts are rendered with an auxiliary decoder for visualization only. Additional rollout visualizations are provided in the Appendix.}
\label{fig:main_visual_rollout}
\end{figure}

\section{Experiments}
\label{sec:experiments}

\subsection{Experimental Setup}
\label{sec:exp_setup}

\begin{table*}[t]
\centering
\caption{Normalized MSE and Pearson correlation $r$ results on AntMaze-L. Lower MSE is better, and higher $r$ is better. Best results are bolded; second-best and third-best cells under each metric are shaded in light red and light blue, respectively.}
\label{tab:antmaze_results}
\small
\begin{tabular}{lcccc}
\toprule
\multirow{2}{*}{\textbf{Method}}
& \multicolumn{2}{c}{\textbf{Linear}}
& \multicolumn{2}{c}{\textbf{MLP}} \\
\cmidrule(lr){2-3}
\cmidrule(lr){4-5}
& \textbf{MSE}$\downarrow$
& \textbf{$r$}$\uparrow$
& \textbf{MSE}$\downarrow$
& \textbf{$r$}$\uparrow$ \\
\midrule

LeWM~\cite{lewm}
& \secondrank{0.901\pmstd{0.248}}
& \secondrank{0.199\pmstd{0.250}}
& \secondrank{0.804\pmstd{0.294}}
& \secondrank{0.328\pmstd{0.309}} \\

C-JEPA~\cite{c-jepa}
& 0.925\pmstd{0.244}
& 0.175\pmstd{0.235}
& 0.830\pmstd{0.290}
& 0.305\pmstd{0.305} \\

GCIQL~\cite{ogbench,kostrikov2022iql}
& \thirdrank{0.908\pmstd{0.245}}
& \thirdrank{0.190\pmstd{0.242}}
& \thirdrank{0.812\pmstd{0.290}}
& \thirdrank{0.320\pmstd{0.305}} \\

GCIVL~\cite{ogbench}
& 0.930\pmstd{0.255}
& 0.170\pmstd{0.230}
& 0.835\pmstd{0.298}
& 0.300\pmstd{0.310} \\

PLDM~\cite{PLDM}
& 0.922\pmstd{0.250}
& 0.180\pmstd{0.235}
& 0.825\pmstd{0.295}
& 0.310\pmstd{0.302} \\

EB-JEPA~\cite{EB-JEPA}
& 0.915\pmstd{0.240}
& 0.185\pmstd{0.240}
& 0.820\pmstd{0.285}
& 0.315\pmstd{0.300} \\

TD-MPC2~\cite{TD-MPC2}
& 0.945\pmstd{0.260}
& 0.155\pmstd{0.225}
& 0.850\pmstd{0.305}
& 0.285\pmstd{0.315} \\

Sub-JEPA~\cite{Sub-JEPA}
& 0.920\pmstd{0.242}
& 0.182\pmstd{0.238}
& 0.823\pmstd{0.292}
& 0.312\pmstd{0.301} \\

GeoWorld-Style~\cite{zhang2026geoworld} 
& 0.912\pmstd{0.248} 
& 0.188\pmstd{0.245} 
& 0.816\pmstd{0.290} 
& 0.316\pmstd{0.305} \\

Random
& 0.995\pmstd{0.280}
& 0.050\pmstd{0.150}
& 0.985\pmstd{0.320}
& 0.065\pmstd{0.200} \\

\midrule
\rowcolor[HTML]{E6E6E6}
\textbf{ProWorld (Ours)}
& \textbf{0.900}\pmstdb{0.246}
& \textbf{0.205}\pmstdb{0.248}
& \textbf{0.801}\pmstdb{0.296}
& \textbf{0.331}\pmstdb{0.309} \\

\bottomrule
\end{tabular}
\end{table*}

\begin{table*}[!ht]
    \centering
    \begin{minipage}[t]{0.48\textwidth}
        \vspace{0pt}
        \centering
        \caption{Ablation study on the entailment cone design on Cube-S. Blank entries indicate the same setting as the full model; \xmark indicates removal or replacement of the corresponding component. Best SR is bolded; second-best and third-best SR cells are shaded in light red and light blue, respectively.}
        \label{tab:cone_design_ablation}
        \vspace{1mm}
        \small
        \resizebox{\linewidth}{!}{
        \begin{tabular}{lccccr}
            \toprule
            \textbf{Variant} & \textbf{Angle} & \textbf{Dist.} & \textbf{Adapt.} & \textbf{Order} & \textbf{SR} \\
            \midrule
            \rowcolor[HTML]{e6e6e6} \textbf{Full}             &  &  &  &  & \textbf{78\%} \\
            Seg. Cone        &  &  &  &  & \secondrank{76\%} \\
            Ang. Only        &  & \xmark &  &  & 66\% \\
            Dist. Only       & \xmark &  &  &  & 62\% \\
            Fixed            &  &  & \xmark &  & \thirdrank{70\%} \\
            Rev.             &  &  &  & \xmark & 44\% \\
            \bottomrule
        \end{tabular}
        }
    \end{minipage}
    \hfill
    \begin{minipage}[t]{0.48\textwidth}
        \vspace{0pt}
        \centering
        \caption{Ablation study on planning cost terms on Cube-S. Blank entries indicate the full planning cost is retained; \xmark indicates removal of the corresponding cost term. Best SR is bolded; second-best and third-best SR cells are shaded in light red and light blue, respectively.}
        \label{tab:planning_cost_ablation}
        \vspace{1mm}
        \footnotesize
        \resizebox{\linewidth}{!}{
        \begin{tabular}{lcccr}
        \toprule
        \textbf{Variant} & \textbf{$D_L^g$} & $C_{\mathrm{best}}$ & $C_{\mathrm{mean}}$ & SR \\
        \midrule
        \rowcolor[HTML]{e6e6e6} \textbf{Full planner} &  &  &  & \textbf{78\%} \\
        Terminal only  &  & \xmark & \xmark & 62\% \\
        w/o best       &  & \xmark &  & \thirdrank{64\%} \\
        w/o mean       &  &  & \xmark & \secondrank{72\%} \\
        \bottomrule
        \end{tabular}
        }
    \end{minipage}
\end{table*}

\subsubsection{Datasets}
\label{sec:tasks_datasets}

We evaluate on four tasks as our main benchmarks: \textbf{PushT}, \textbf{OGBench-Cube-Single-Play}, \textbf{OGBench-AntMaze-Large-Navigate}, and \textbf{OGBench-Scene-Play}. For brevity, we refer to the latter three tasks as \textbf{Cube-S}, \textbf{AntMaze-L}, and \textbf{Scene}, respectively. This suite covers complementary planning scenarios, including contact-rich planar pushing, 3D object manipulation, long-horizon visual navigation, and mixed discrete-continuous multi-object scene manipulation. \textbf{Further dataset details are provided in the Appendix.}

\begin{figure}[t]
\centering
\includegraphics[width=\linewidth]{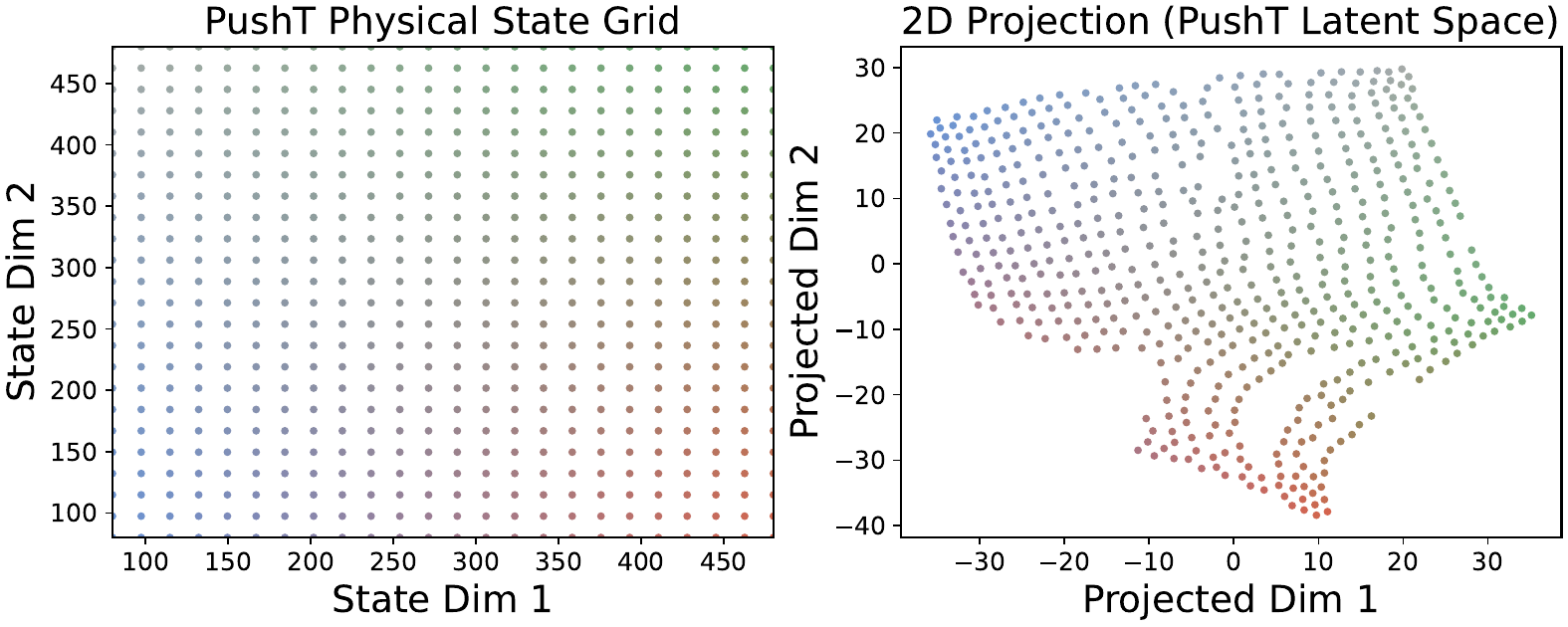}
\caption{PushT physical state grid and t-SNE projection of ProWorld latents. The visualization shows local continuity and a radial expansion trend in the learned latent space.}
\label{fig:pusht_tsne}
\end{figure}

\subsubsection{Implementation Details}
\label{sec:implementation_details}
All experiments are conducted on a Huawei Ascend 910C NPU. The complete hyperparameter settings are provided in the Appendix.

\subsubsection{Comparison with SOTA Methods}
\label{sec:comparison}

We compare ProWorld against the following recent baselines: LeWM~\cite{lewm}, C-JEPA~\cite{c-jepa}, GCIQL~\cite{ogbench,kostrikov2022iql}, GCIVL~\cite{ogbench}, PLDM~\cite{PLDM}, EB-JEPA~\cite{EB-JEPA}, TD-MPC2~\cite{TD-MPC2}, and Sub-JEPA~\cite{Sub-JEPA}. All methods use identical training data and planning budgets to ensure a fair comparison.

Tab.~\ref{tab:main_results} summarizes the success rates of ProWorld and the baselines across different tasks. Compared to LeWM, ProWorld improves SR by an average absolute margin of approximately 9.67 across the four main tasks. This result suggests that world models relying solely on local next-step prediction remain insufficient for long-horizon goal-conditioned tasks, whereas explicitly modeling progress-hierarchical relationships improves long-term planning capability.

A closer look at task categories shows that ProWorld yields larger gains on long-horizon tasks, especially AntMaze-L, than on short-horizon 2D manipulation tasks. This suggests that local transition prediction is often sufficient when goals are reachable within a few steps, but becomes less reliable when planning requires sustained progress across many intermediate states.

\subsubsection{Case Study}
To further analyze ProWorld's behavior in long-horizon prediction, Fig.~\ref{fig:main_visual_rollout} presents a representative open-loop rollout on the PushT task. Given the first few context frames, the model must recursively predict subsequent states without receiving future ground-truth observations. As shown, LeWM's predictions maintain reasonable local appearance but, as the number of rollout steps increases, the predicted object poses and goal-relevant positions progressively diverge from the ground-truth trajectory. In contrast, ProWorld's predicted trajectory remains closer to the ground truth over a longer time horizon and more consistently maintains the trend of advancing the object toward the goal state.

\begin{figure}[t]
\centering
\includegraphics[width=\linewidth]{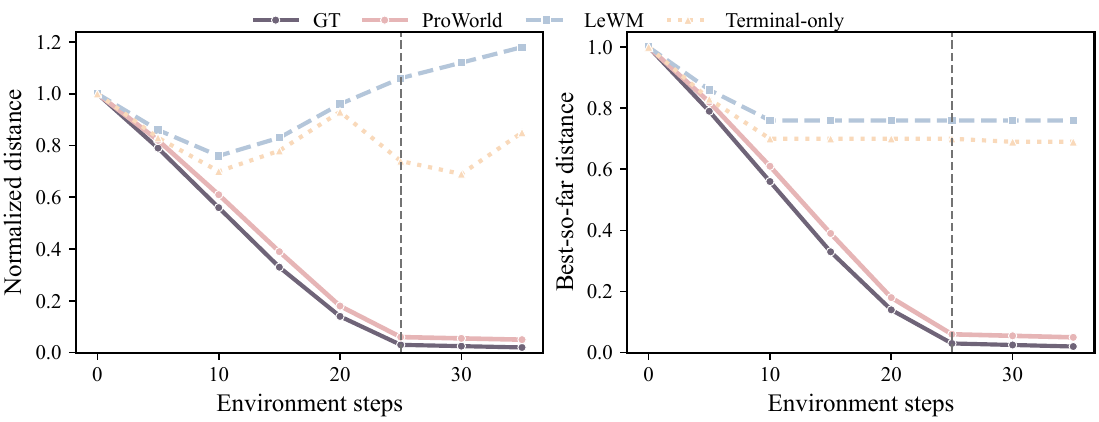}
\caption{Planning-time progress on PushT. ProWorld maintains a goal-distance profile closer to ground-truth progress than LeWM and terminal-only planning.}
\label{fig:planning_case_progress}
\end{figure}

Fig.~\ref{fig:planning_case_progress} further extends the representation visualization in Fig.~\ref{fig:pusht_tsne} from latent organization to planning-time decision behavior. Here, normalized distance denotes the current goal distance divided by the initial goal distance, while best-so-far distance denotes the running minimum of this normalized distance up to the current environment step. Although LeWM and the terminal-only planner can produce locally plausible candidate rollouts, their goal distances stagnate or increase over intermediate steps. By contrast, ProWorld selects a rollout whose distance profile remains closer to the ground-truth progress curve, showing that the intermediate progress terms in the planning cost help reject candidates that appear reasonable only at a local or terminal level.

\subsection{Ablation Study}
\label{sec:ablation}
To analyze the sources of ProWorld's performance, we conduct ablation studies on Cube-S. Blank entries in the table indicate the same setting as the full model; \xmark indicates removal or replacement of the corresponding component. Best, second-best, and third-best results are bolded, shaded in light red, and shaded in light blue, respectively. \textbf{Additional ablation results, visualizations, and table-abbreviation details are provided in the Appendix.}

% \begin{figure*}[t]
%     \centering
%     \begin{subfigure}[t]{0.40\textwidth}
%         \centering
%         \includegraphics[width=\linewidth]{figures/appendix_spearman.pdf}
%         \caption{Global progress correlation.}
%         \label{fig:progress_correlation}
%     \end{subfigure}
%     \hfill
%     \begin{minipage}[t]{0.56\textwidth}
%         \centering
%         \begin{subfigure}[t]{0.48\linewidth}
%             \centering
%             \includegraphics[width=\linewidth]{figures/parent_child.pdf}
%             \caption{Radial change.}
%             \label{fig:parent_child_radius_gain}
%         \end{subfigure}
%         \hfill
%         \begin{subfigure}[t]{0.48\linewidth}
%             \centering
%             \includegraphics[width=\linewidth]{figures/parent_child_ordered_radius.pdf}
%             \caption{Radius ordering.}
%             \label{fig:parent_child_ordered_radius}
%         \end{subfigure}
%     \end{minipage}
%     \caption{Analysis of the progress geometry learned by ProWorld on PushT. (a) Normalized trajectory progress is positively correlated with Lorentz radius and more strongly correlated with goal-direction alignment. (b) Temporally ordered child states generally exhibit positive radial displacement from their parents. (c) A majority of sampled pairs place the child at a larger radius. Together, these results support a joint radial and directional organization that is statistical rather than strictly monotonic.}
%     \label{fig:progress_geometry_analysis}
% \end{figure*}

\paragraph{Learning objectives.}
Tab.~\ref{tab:ablation} evaluates the contribution of each training objective in ProWorld, where Pred., Ctr., Cone., and Reg. denote $\mathcal{L}_{\mathrm{pred}}$, $\mathcal{L}_{\mathrm{ctr}}$, $\mathcal{L}_{\mathrm{cone}}$, and $\mathcal{L}_{\mathrm{sig}}$, respectively. The full model achieves 78\% SR, substantially outperforming the Pred.-only baseline at 46\%, showing that local latent prediction alone is insufficient for long-horizon visual goal planning. Removing the contrastive loss causes the largest drop, nearly reducing performance to the Pred. baseline, which indicates that future-state discriminability is essential. Removing the entailment cone also degrades SR to 62\%, confirming the importance of directional progress structure, while removing SIGReg has only a minor effect, suggesting that it mainly stabilizes representation learning.

\paragraph{Progress geometry.}

\begin{figure}[!t]
\centering
\includegraphics[width=0.96\linewidth]{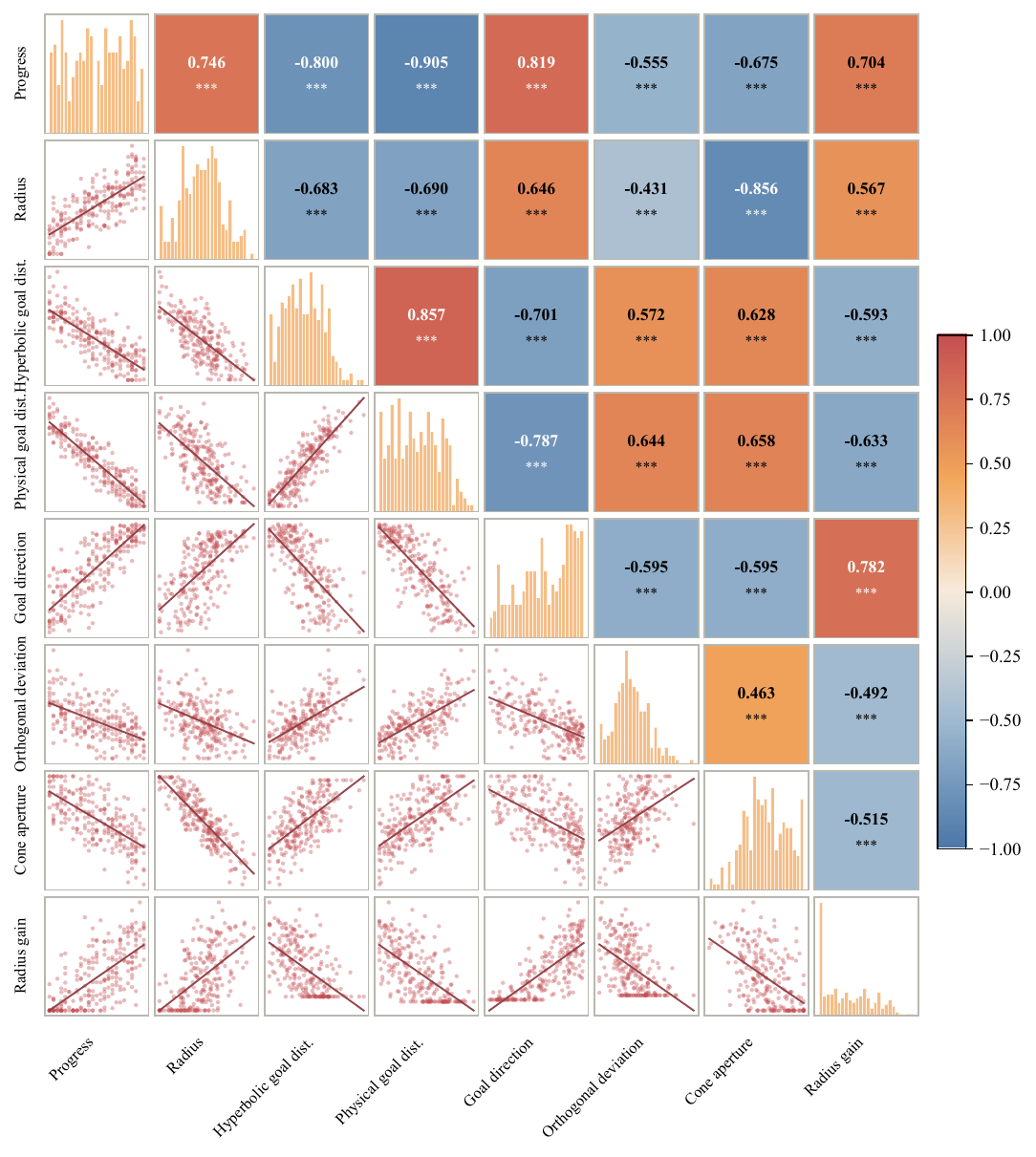}
\caption{Spearman correlation analysis of progress-related geometric variables on PushT.}
\label{fig:pusht_progress_correlation}
\end{figure}

\begin{figure}[!t]
\centering
\begin{minipage}[t]{0.48\textwidth}
    \centering
    \includegraphics[width=\linewidth]{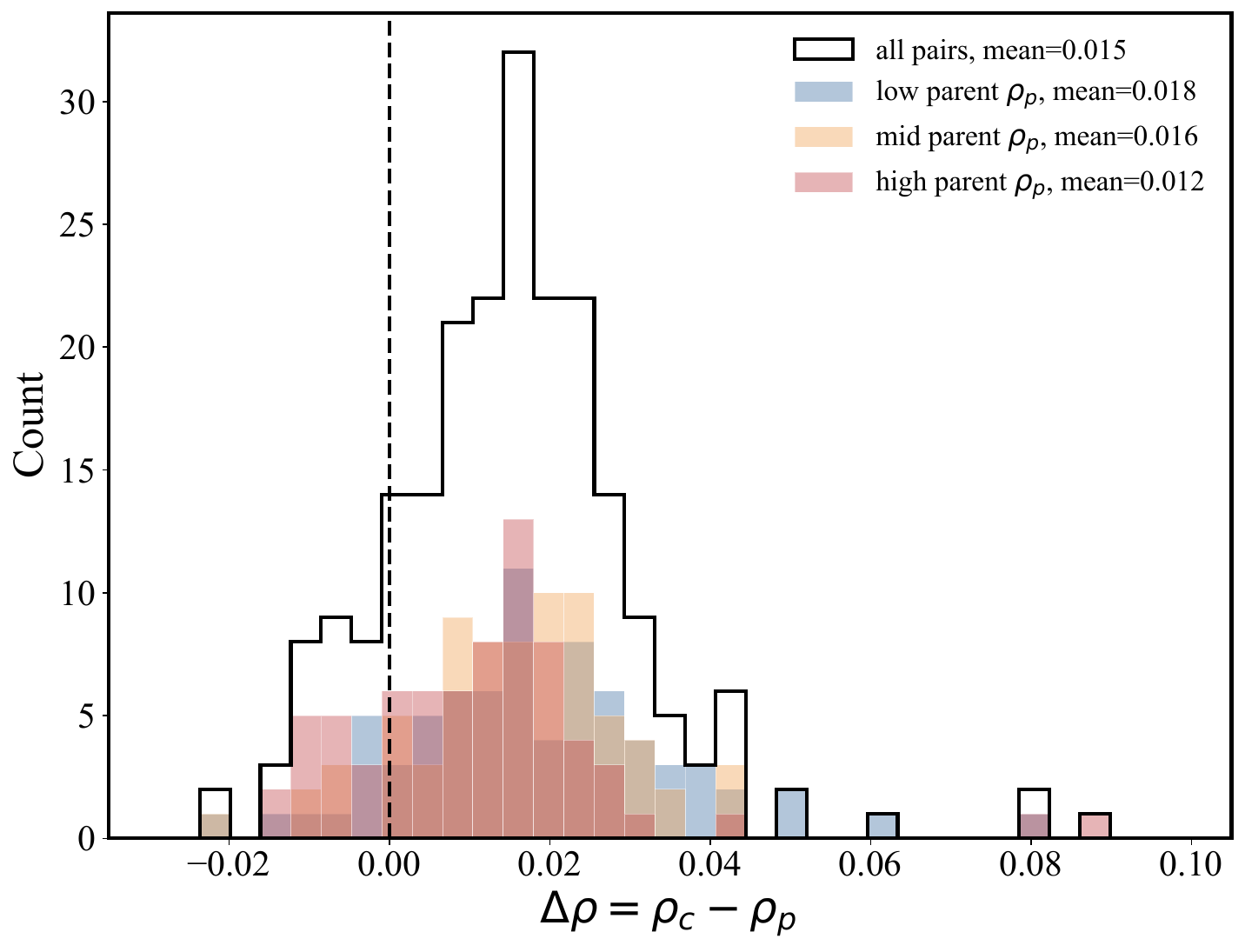}
    \centerline{(a) PushT}
\end{minipage}
\hfill
\begin{minipage}[t]{0.48\textwidth}
    \centering
    \includegraphics[width=\linewidth]{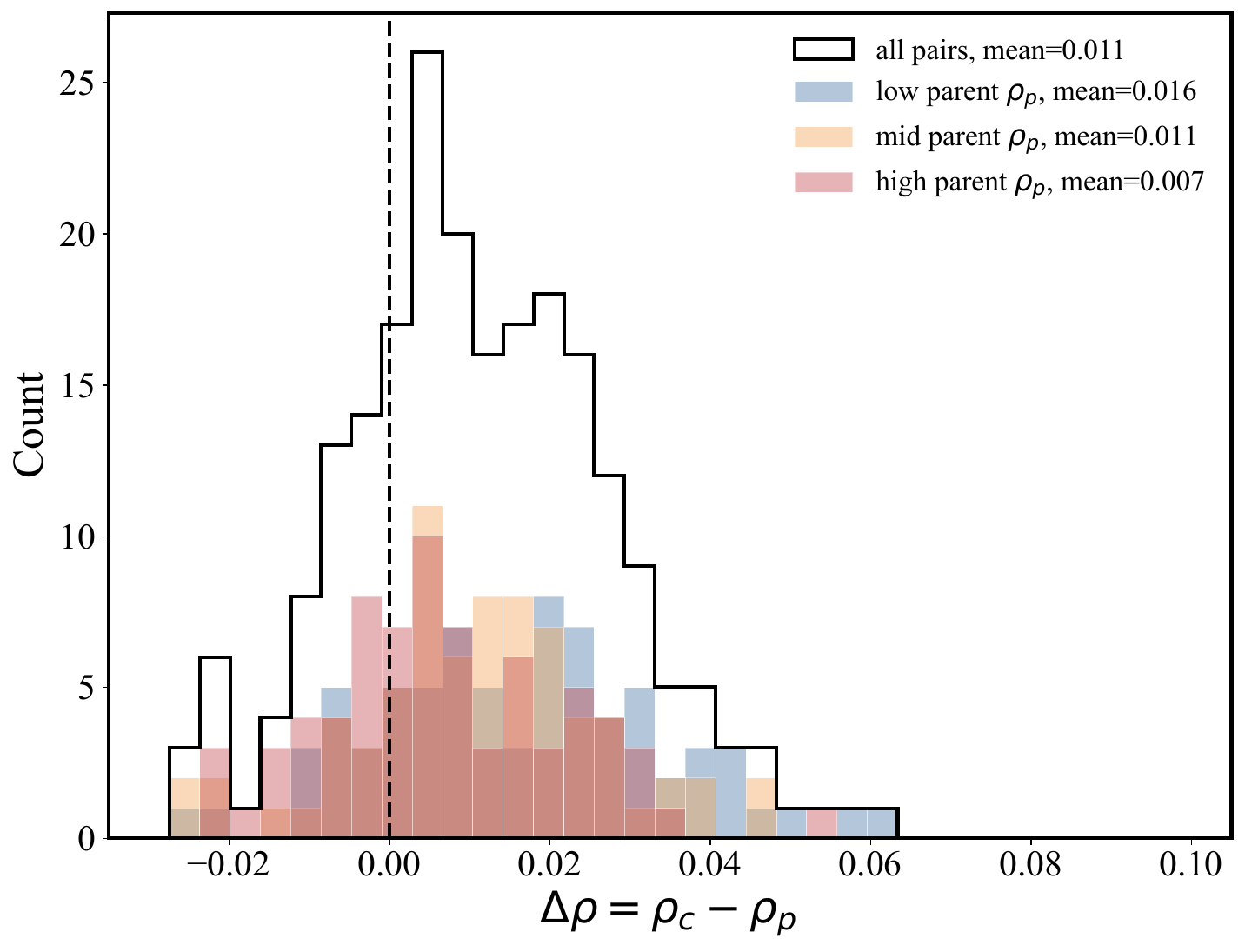}
    \centerline{(b) Cube-S}
\end{minipage}
\caption{Diagnostic parent-child radial displacement grouped by parent radius. Positive values indicate that the child has a larger Lorentz radius than the parent.}
\label{fig:parent_radius_grouped_gain}
\end{figure}

Figs.~\ref{fig:pusht_tsne}, \ref{fig:pusht_progress_correlation}, and~\ref{fig:parent_radius_grouped_gain} analyze the learned progress structure. The t-SNE visualization shows that ProWorld preserves local continuity while forming a structured latent space. The correlation diagnostic further shows that trajectory progress is positively associated with goal-direction alignment and negatively associated with both hyperbolic and physical goal distances, consistent with the goal-anchored entailment objective. The radial diagnostics additionally show an emergent coarse-to-fine organization, where later states tend to occupy more specific regions of the Lorentz space.

\paragraph{Geometry and progress order.}
Tab.~\ref{tab:geometry_order_ablation} examines the role of geometry and state-pair construction. Replacing the Lorentz hyperbolic space with a Euclidean latent space reduces SR from 78\% to 68\%, suggesting that negative curvature better supports coarse-to-fine progress organization. More importantly, reversing the parent-child order or using random state pairs drops SR to 44\% and 52\%, respectively, showing that the entailment cone relies on correct goal-conditioned temporal order rather than acting as a generic regularizer. The fixed-aperture variant reaches 70\% SR, further indicating that adaptive cone apertures better match goal-conditioned progress, where states farther from the goal allow broader future regions and states closer to the goal impose finer directional constraints.

\paragraph{Entailment cone design.}
Tab.~\ref{tab:cone_design_ablation} decomposes the entailment cone design. Using only the angular or goal-distance term performs clearly worse than the full model, showing that directional consistency and distance-based progress are complementary. The segment-cone variant that additionally constrains $h_{t+1}$ reaches 76\% SR, close to but below the full model, suggesting that the default $\Delta=2$ pair provides sufficient progress supervision without over-constraining adjacent transitions. Fixed apertures and reversed temporal order further degrade performance, confirming the importance of adaptive and correctly ordered goal-conditioned cones.

\paragraph{Planning objective.}
Tab.~\ref{tab:planning_cost_ablation} evaluates the progress-aware planning cost. Using only the terminal distance reduces SR from 78\% to 62\%, showing that endpoint proximity alone is insufficient for selecting reliable long-horizon rollouts. Removing either the best-intermediate or mean-distance term also degrades performance, indicating that intermediate states provide important progress signals for rejecting locally plausible but globally ineffective action sequences.

\section{Conclusion}
\label{sec:conclusion}

We presented ProWorld, a progress-aware hyperbolic world model for visual goal planning. It models goal-conditioned progress order in Lorentz space to improve future discrimination, directional progress learning, and rollout scoring. Experiments on four tasks show consistent gains over strong baselines.

{
    \bibliography{aaai2026}

@inproceedings{finn2017deep,
  title     = {Deep Visual Foresight for Planning Robot Motion},
  author    = {Finn, Chelsea and Levine, Sergey},
  booktitle = {2017 IEEE International Conference on Robotics and Automation (ICRA)},
  pages     = {2786--2793},
  year      = {2017},
  organization = {IEEE}
}

@inproceedings{hafner2019planet,
  title     = {Learning Latent Dynamics for Planning from Pixels},
  author    = {Hafner, Danijar and Lillicrap, Timothy and Fischer, Ian and Villegas, Ruben and Ha, David and Lee, Honglak and Davidson, James},
  booktitle = {Proceedings of the 36th International Conference on Machine Learning},
  pages     = {2555--2565},
  year      = {2019},
  series    = {Proceedings of Machine Learning Research},
  volume    = {97},
  publisher = {PMLR}
}

@inproceedings{ha2018worldmodels,
  title     = {Recurrent World Models Facilitate Policy Evolution},
  author    = {Ha, David and Schmidhuber, J{\"u}rgen},
  booktitle = {Advances in Neural Information Processing Systems},
  volume    = {31},
  year      = {2018}
}

@article{lecun2022path,
  title={A path towards autonomous machine intelligence version 0.9. 2, 2022-06-27},
  author={LeCun, Yann and others},
  journal={Open Review},
  volume={62},
  number={1},
  pages={1--62},
  year={2022}
}

@inproceedings{assran2023ijepa,
  title     = {Self-Supervised Learning from Images with a Joint-Embedding Predictive Architecture},
  author    = {Assran, Mahmoud and Duval, Quentin and Misra, Ishan and Bojanowski, Piotr and Vincent, Pascal and Rabbat, Michael and LeCun, Yann and Ballas, Nicolas},
  booktitle = {Proceedings of the IEEE/CVF Conference on Computer Vision and Pattern Recognition},
  pages     = {15619--15629},
  year      = {2023}
}

@misc{balestriero2025lejepa,
  title         = {LeJEPA: Provable and Scalable Self-Supervised Learning Without the Heuristics},
  author        = {Balestriero, Randall and LeCun, Yann},
  year          = {2025},
  eprint        = {2511.08544},
  archivePrefix = {arXiv},
  primaryClass  = {cs.LG}
}

@misc{lewm,
  title         = {LeWorldModel: Stable End-to-End Joint-Embedding Predictive Architecture from Pixels},
  author        = {Maes, Lucas and Le Lidec, Quentin and Scieur, Damien and LeCun, Yann and Balestriero, Randall},
  year          = {2026},
  eprint        = {2603.19312},
  archivePrefix = {arXiv},
  primaryClass  = {cs.LG}
}

@inproceedings{ogbench,
  title     = {OGBench: Benchmarking Offline Goal-Conditioned RL},
  author    = {Park, Seohong and Frans, Kevin and Eysenbach, Benjamin and Levine, Sergey},
  booktitle = {International Conference on Learning Representations},
  year      = {2025}
}

@inproceedings{nickel2018lorentz,
  title     = {Learning Continuous Hierarchies in the Lorentz Model of Hyperbolic Geometry},
  author    = {Nickel, Maximilian and Kiela, Douwe},
  booktitle = {Proceedings of the 35th International Conference on Machine Learning},
  pages     = {3779--3788},
  year      = {2018},
  series    = {Proceedings of Machine Learning Research},
  volume    = {80},
  publisher = {PMLR}
}

@inproceedings{ganea2018hyperbolic,
  title     = {Hyperbolic Entailment Cones for Learning Hierarchical Embeddings},
  author    = {Ganea, Octavian-Eugen and B{\'e}cigneul, Gary and Hofmann, Thomas},
  booktitle = {Proceedings of the 35th International Conference on Machine Learning},
  pages     = {1646--1655},
  year      = {2018},
  series    = {Proceedings of Machine Learning Research},
  volume    = {80},
  publisher = {PMLR}
}

@inproceedings{hafner2020dreamer,
  title     = {Dream to Control: Learning Behaviors by Latent Imagination},
  author    = {Hafner, Danijar and Lillicrap, Timothy and Ba, Jimmy and Norouzi, Mohammad},
  booktitle = {International Conference on Learning Representations},
  year      = {2020}
}

@misc{zhang2026geoworld,
  title         = {GeoWorld: Geometric World Models},
  author        = {Zhang, Zeyu and Li, Danning and Reid, Ian and Hartley, Richard},
  year          = {2026},
  eprint        = {2602.23058},
  archivePrefix = {arXiv},
  primaryClass  = {cs.LG},
  url           = {https://arxiv.org/abs/2602.23058}
}

@article{c-jepa,
  title={Causal-JEPA: Learning World Models through Object-Level Latent Masking},
  author={Nam, Heejeong and Le Lidec, Quentin and Maes, Lucas and LeCun, Yann and Balestriero, Randall},
  journal={arXiv preprint arXiv:2602.11389},
  year={2026},
  url={https://arxiv.org/abs/2602.11389}
}

@inproceedings{kostrikov2022iql,
  title={Offline Reinforcement Learning with Implicit Q-Learning},
  author={Kostrikov, Ilya and Nair, Ashvin and Levine, Sergey},
  booktitle={International Conference on Learning Representations},
  year={2022},
  url={https://openreview.net/forum?id=68n2s9ZJWF8}
}

@misc{EB-JEPA,
      title={A Lightweight Library for Energy-Based Joint-Embedding Predictive Architectures},
      author={Basile Terver and Randall Balestriero and Megi Dervishi and David Fan and Quentin Garrido and Tushar Nagarajan and Koustuv Sinha and Wancong Zhang and Mike Rabbat and Yann LeCun and Amir Bar},
      year={2026},
      eprint={2602.03604},
      archivePrefix={arXiv},
      primaryClass={cs.CV},
      url={https://arxiv.org/abs/2602.03604},
}

@inproceedings{TD-MPC2,
  title={{TD-MPC2}: Scalable, Robust World Models for Continuous Control},
  author={Hansen, Nicklas and Su, Hao and Wang, Xiaolong},
  booktitle={International Conference on Learning Representations},
  year={2024}
}

@inproceedings{Dino-wm,
  title={{DINO-WM}: World Models on Pre-trained Visual Features Enable Zero-shot Planning},
  author={Zhou, Gaoyue and Pan, Hengkai and LeCun, Yann and Pinto, Lerrel},
  booktitle={Proceedings of the 42nd International Conference on Machine Learning},
  year={2025},
  url={https://openreview.net/forum?id=D5RNACOZEI}
}

@misc{Sub-JEPA,
  title        = {Sub-JEPA: Subspace Gaussian Regularization for Stable End-to-End World Models},
  author       = {Zhao, Kai and Nie, Dongliang and Lin, Yuchen and Luo, Zhehan and Gu, Yixiao and Fan, Deng-Ping and Zeng, Dan},
  year         = {2026},
  eprint       = {2605.09241},
  archivePrefix = {arXiv},
  primaryClass = {cs.LG},
  url          = {https://arxiv.org/abs/2605.09241}
}

@inproceedings{PLDM,
  title={Stress-Testing Offline Reward-Free Reinforcement Learning: A Case for Planning with Latent Dynamics Models},
  author={Sobal, Vlad and Zhang, Wancong and Cho, Kyunghyun and Balestriero, Randall and Rudner, Tim G. J. and LeCun, Yann},
  booktitle={ICLR Workshop on World Models for Robot Learning},
  year={2025},
  url={https://openreview.net/forum?id=jON7H6A9UU}
}

@inproceedings{watter2015embed,
  title     = {Embed to Control: A Locally Linear Latent Dynamics Model for Control from Raw Images},
  author    = {Watter, Manuel and Springenberg, Jost Tobias and Boedecker, Joschka and Riedmiller, Martin},
  booktitle = {Advances in Neural Information Processing Systems},
  volume    = {28},
  year      = {2015}
}

@inproceedings{pertsch2020longhorizon,
  title     = {Long-Horizon Visual Planning with Goal-Conditioned Hierarchical Predictors},
  author    = {Pertsch, Karl and Rybkin, Oleh and Ebert, Frederik and Zhou, Shenghao and Jayaraman, Dinesh and Finn, Chelsea and Levine, Sergey},
  booktitle = {Advances in Neural Information Processing Systems},
  volume    = {33},
  pages     = {17321--17333},
  year      = {2020}
}

@inproceedings{janner2019mbpo,
  title     = {When to Trust Your Model: Model-Based Policy Optimization},
  author    = {Janner, Michael and Fu, Justin and Zhang, Marvin and Levine, Sergey},
  booktitle = {Advances in Neural Information Processing Systems},
  volume    = {32},
  year      = {2019}
}

@inproceedings{lambert2020objective,
  title     = {Objective Mismatch in Model-Based Reinforcement Learning},
  author    = {Lambert, Nathan and Amos, Brandon and Yadan, Omry and Calandra, Roberto},
  booktitle = {Proceedings of the 2nd Conference on Learning for Dynamics and Control},
  pages     = {761--770},
  year      = {2020},
  series    = {Proceedings of Machine Learning Research},
  volume    = {120},
  publisher = {PMLR}
}

@inproceedings{nair2020goalaware,
  title     = {Goal-Aware Prediction: Learning to Model What Matters},
  author    = {Nair, Suraj and Savarese, Silvio and Finn, Chelsea},
  booktitle = {Proceedings of the 37th International Conference on Machine Learning},
  pages     = {7207--7219},
  year      = {2020},
  series    = {Proceedings of Machine Learning Research},
  volume    = {119},
  publisher = {PMLR}
}

@inproceedings{gieselmann2023expansive,
  title     = {Expansive Latent Planning for Sparse Reward Offline Reinforcement Learning},
  author    = {Gieselmann, Robert and Pokorny, Florian T.},
  booktitle = {Proceedings of the 7th Conference on Robot Learning},
  pages     = {1--22},
  year      = {2023},
  series    = {Proceedings of Machine Learning Research},
  volume    = {229},
  publisher = {PMLR}
}

@inproceedings{eysenbach2022contrastive,
  title     = {Contrastive Learning as Goal-Conditioned Reinforcement Learning},
  author    = {Eysenbach, Benjamin and Zhang, Tianjun and Levine, Sergey and Salakhutdinov, Ruslan},
  booktitle = {Advances in Neural Information Processing Systems},
  volume    = {35},
  year      = {2022}
}

@inproceedings{khrulkov2020hyperbolic,
  title     = {Hyperbolic Image Embeddings},
  author    = {Khrulkov, Valentin and Mirvakhabova, Leyla and Ustinova, Evgeniya and Oseledets, Ivan and Lempitsky, Victor},
  booktitle = {Proceedings of the IEEE/CVF Conference on Computer Vision and Pattern Recognition},
  pages     = {6418--6428},
  year      = {2020}
}

@inproceedings{ermolov2022hyperbolic,
  title     = {Hyperbolic Vision Transformers: Combining Improvements in Metric Learning},
  author    = {Ermolov, Aleksandr and Mirvakhabova, Leyla and Khrulkov, Valentin and Sebe, Nicu and Oseledets, Ivan},
  booktitle = {Proceedings of the IEEE/CVF Conference on Computer Vision and Pattern Recognition},
  pages     = {7409--7419},
  year      = {2022}
}

@article{schrittwieser2020muzero,
  title   = {Mastering Atari, Go, Chess and Shogi by Planning with a Learned Model},
  author  = {Schrittwieser, Julian and Antonoglou, Ioannis and Hubert, Thomas and Simonyan, Karen and Sifre, Laurent and Schmitt, Simon and Guez, Arthur and Lockhart, Edward and Hassabis, Demis and Graepel, Thore and Lillicrap, Timothy and Silver, David},
  journal = {Nature},
  volume  = {588},
  pages   = {604--609},
  year    = {2020}
}

@inproceedings{hansen2022tdmpc,
  title     = {Temporal Difference Learning for Model Predictive Control},
  author    = {Hansen, Nicklas A. and Su, Hao and Wang, Xiaolong},
  booktitle = {Proceedings of the 39th International Conference on Machine Learning},
  pages     = {8387--8406},
  year      = {2022},
  series    = {Proceedings of Machine Learning Research},
  volume    = {162},
  publisher = {PMLR}
}

@article{bardes2024vjepa,
  title   = {Revisiting Feature Prediction for Learning Visual Representations from Video},
  author  = {Bardes, Adrien and Garrido, Quentin and Ponce, Jean and Chen, Xinlei and Rabbat, Michael and LeCun, Yann and Assran, Mido and Ballas, Nicolas},
  journal = {Transactions on Machine Learning Research},
  year    = {2024},
  url     = {https://openreview.net/forum?id=QaCCuDfBk2}
}

@inproceedings{eysenbach2019sorb,
  title     = {Search on the Replay Buffer: Bridging Planning and Reinforcement Learning},
  author    = {Eysenbach, Benjamin and Salakhutdinov, Ruslan and Levine, Sergey},
  booktitle = {Advances in Neural Information Processing Systems},
  volume    = {32},
  year      = {2019}
}

@inproceedings{liu2020hyperbolicvisual,
  title     = {Hyperbolic Visual Embedding Learning for Zero-Shot Recognition},
  author    = {Liu, Shaoteng and Chen, Jingjing and Pan, Liangming and Ngo, Chong-Wah and Chua, Tat-Seng and Jiang, Yu-Gang},
  booktitle = {Proceedings of the IEEE/CVF Conference on Computer Vision and Pattern Recognition},
  pages     = {9273--9281},
  year      = {2020}
}

@inproceedings{atigh2022hyperbolicseg,
  title     = {Hyperbolic Image Segmentation},
  author    = {Atigh, Mina Ghadimi and Schoep, Julian and Acar, Erman and van Noord, Nanne and Mettes, Pascal},
  booktitle = {Proceedings of the IEEE/CVF Conference on Computer Vision and Pattern Recognition},
  pages     = {4453--4462},
  year      = {2022}
}

@inproceedings{desai2023meru,
  title     = {Hyperbolic Image-Text Representations},
  author    = {Desai, Karan and Nickel, Maximilian and Rajpurohit, Tanmay and Johnson, Justin and Vedantam, Shanmukha Ramakrishna},
  booktitle = {Proceedings of the 40th International Conference on Machine Learning},
  pages     = {7694--7731},
  year      = {2023},
  series    = {Proceedings of Machine Learning Research},
  volume    = {202},
  publisher = {PMLR}
}

@inproceedings{ge2023hyperboliccontrastive,
  title     = {Hyperbolic Contrastive Learning for Visual Representations Beyond Objects},
  author    = {Ge, Songwei and Mishra, Shlok and Kornblith, Simon and Li, Chun-Liang and Jacobs, David},
  booktitle = {Proceedings of the IEEE/CVF Conference on Computer Vision and Pattern Recognition},
  pages     = {6840--6849},
  year      = {2023}
}

@misc{bui2025arrow,
  title         = {Learning Along the Arrow of Time: Hyperbolic Geometry for Backward-Compatible Representation Learning},
  author        = {Bui, Ngoc and Yang, Menglin and Chen, Runjin and Neves, Leonardo and Ju, Mingxuan and Ying, Rex and Shah, Neil and Zhao, Tong},
  year          = {2025},
  eprint        = {2506.05826},
  archivePrefix = {arXiv},
  primaryClass  = {cs.LG},
  url           = {https://arxiv.org/abs/2506.05826}
}

@misc{madhu2026hyprag,
  title         = {HypRAG: Hyperbolic Dense Retrieval for Retrieval Augmented Generation},
  author        = {Madhu, Hiren and Bui, Ngoc and Maatouk, Ali and Tassiulas, Leandros and Krishnaswamy, Smita and Yang, Menglin and Ganguly, Sukanta and Srinivasan, Kiran and Ying, Rex},
  year          = {2026},
  eprint        = {2602.07739},
  archivePrefix = {arXiv},
  primaryClass  = {cs.IR},
  url           = {https://arxiv.org/abs/2602.07739}
}

@misc{sohn2022bending,
  title         = {Bending the Future: Autoregressive Modeling of Temporal Knowledge Graphs in Curvature-Variable Hyperbolic Spaces},
  author        = {Sohn, Jihoon and Ma, Mingyu Derek and Chen, Muhao},
  year          = {2022},
  eprint        = {2209.05635},
  archivePrefix = {arXiv},
  primaryClass  = {cs.LG},
  url           = {https://arxiv.org/abs/2209.05635}
}

@misc{zhang2025hmamba,
  title         = {HMamba: Hyperbolic Mamba for Sequential Recommendation},
  author        = {Zhang, Qianru and Wen, Honggang and Yuan, Wei and Chen, Crystal and Yang, Menglin and Yiu, Siu-Ming and Yin, Hongzhi},
  year          = {2025},
  eprint        = {2505.09205},
  archivePrefix = {arXiv},
  primaryClass  = {cs.IR},
  url           = {https://arxiv.org/abs/2505.09205}
}

@misc{cetin2022hyperbolicrl,
  title         = {Hyperbolic Deep Reinforcement Learning},
  author        = {Cetin, Edoardo and Chamberlain, Benjamin and Bronstein, Michael and Hunt, Jonathan J.},
  year          = {2022},
  eprint        = {2210.01542},
  archivePrefix = {arXiv},
  primaryClass  = {cs.LG},
  url           = {https://arxiv.org/abs/2210.01542}
}

@misc{klein2026hyperplusplus,
  title         = {Understanding and Improving Hyperbolic Deep Reinforcement Learning},
  author        = {Klein, Timo and Lang, Thomas and Shkabrii, Andrii and Sturm, Alexander and Sidak, Kevin and Miklautz, Lukas and Plant, Claudia and Velaj, Yllka and Tschiatschek, Sebastian},
  year          = {2026},
  eprint        = {2512.14202},
  archivePrefix = {arXiv},
  primaryClass  = {cs.LG},
  url           = {https://arxiv.org/abs/2512.14202}
}

@misc{liu2026hyperguide,
  title         = {HyperGuide: Hyperbolic Guidance for Efficient Multi-Step Reasoning in Large Language Models},
  author        = {Liu, Yuyu and Xu, Haotian and He, Yanan and Patil, Sarang Rajendra and Xu, Mengjia and Ma, Tengfei},
  year          = {2026},
  eprint        = {2605.24140},
  archivePrefix = {arXiv},
  primaryClass  = {cs.AI},
  url           = {https://arxiv.org/abs/2605.24140}
}
}

\clearpage
\appendix
\setcounter{equation}{0}
\setcounter{section}{0}
\setcounter{proposition}{0}
\setcounter{theorem}{0}

\section{Appendix Overview}

This appendix provides supplementary material that could not be fully elaborated in the main paper due to space constraints.
We first give a compact algorithmic overview, summarize key notation, and provide implementation details for the architecture, hyperparameters, planning settings, datasets, baselines, and evaluation protocol.
We then present a formal discussion of goal-conditioned progress order and report additional analyses, including ablations, sensitivity studies, progress visualization, and qualitative rollouts.

\section{Algorithmic Overview}
\label{sec:algorithmic_overview}

Rather than presenting ProWorld as one monolithic training loop, we decompose the framework into three compact procedure sketches aligned with the main design contributions.
At a high level, ProWorld maps visual observations into a hyperbolic latent space in the Lorentz model, imposes goal-conditioned progress order through goal-anchored entailment constraints, and uses the learned progress geometry to score candidate action rollouts during planning.

\noindent\begin{minipage}{\linewidth}
\paragraph{Innovation I: hyperbolic predictive dynamics.}
This component converts a JEPA-style visual world model into a hyperbolic predictive model with future prediction and future discrimination losses.
\begin{GeoCodeBlock}
def ProWorldDynamics(obs, actions):
    z = encoder(obs)
    h = lorentz_project(z)
    h_pred = hyperbolic_predict(h, actions)

    pred = mean(hyperbolic_dist(h_pred[:, :-1], h[:, 1:]) ** 2)
    ctr = contrastive_future_loss(h_pred, h)

    return pred + lambda_ctr * ctr
\end{GeoCodeBlock}
\end{minipage}
\par\medskip

\noindent\begin{minipage}{\linewidth}
\paragraph{Innovation II: progress-induced entailment.}
This component turns temporally ordered states from goal-conditioned hindsight trajectory segments into weak supervision: later states should lie inside the parent-to-goal cone and move closer to the hindsight goal.
Here \texttt{traj\_id} only keeps sampled triples within the same recorded trajectory, rather than providing a success label.
\begin{GeoCodeBlock}
def ProgressEntailment(h, goal, traj_id, Delta):
    parent, child, h_goal = sample_hindsight_ordered_pair(
        h, goal, traj_id, Delta)
    aperture = adaptive_cone(hyperbolic_dist(parent, h_goal))

    angle = relu(cone_angle(parent, child, h_goal) - aperture)
    progress = relu(hyperbolic_dist(child, h_goal)
                    + m_g - hyperbolic_dist(parent, h_goal))

    return angle + progress
\end{GeoCodeBlock}
\end{minipage}
\par\medskip

\noindent\begin{minipage}{\linewidth}
\paragraph{Innovation III: progress-aware planning.}
This component uses the same hyperbolic geometry at inference time, scoring each CEM rollout by terminal goal proximity and intermediate progress.
\begin{GeoCodeBlock}
def ProgressAwarePlanning(obs, goal, budget):
    h0 = lorentz_project(encoder(obs))
    hg = lorentz_project(encoder(goal))

    for actions in CEM(budget):
        h_roll = hyperbolic_predict(h0, actions)
        D_g = hyperbolic_dist(h_roll, hg) ** 2
        C_T = D_g[-1]
        C_best = min(D_g)
        C_mean = mean(D_g)
        cost = beta_T * C_T + beta_b * C_best + beta_m * C_mean

    return first_action(argmin(cost))
\end{GeoCodeBlock}
\end{minipage}
\par\medskip

\section{Key Notations}

As shown in Tab.~\ref{tab:key_notations}, this section summarizes the key notations used throughout the paper for the reader's reference. For the sake of brevity and readability, we do not present a complete variable table.

\begin{table*}[t]
\centering
\caption{Key notation summary for ProWorld. This table retains only the primary symbols related to hyperbolic geometry, progress relations, entailment cone constraints, and planning objectives.}
\label{tab:key_notations}
\scriptsize
\setlength{\tabcolsep}{4pt}
\renewcommand{\arraystretch}{1.08}
\begin{tabular}{p{0.13\textwidth}p{0.34\textwidth}p{0.13\textwidth}p{0.34\textwidth}}
\toprule
\textbf{Symbol} & \textbf{Description}
& \textbf{Symbol} & \textbf{Description} \\
\midrule
$\mathbb{H}^{m}_{c}$ 
& $m$-dimensional hyperbolic latent space in the Lorentz model with curvature $-c$
& $c$
& Positive curvature parameter of the hyperbolic space, corresponding to a negative spatial curvature of $-c$ \\

$\langle \cdot,\cdot\rangle_{\mathcal{L}}$
& Lorentz inner product
& $d_{\mathbb{H}}(\cdot,\cdot)$
& Hyperbolic geodesic distance \\

$h_t$
& Hyperbolic latent state at time $t$
& $h_g$
& Hyperbolic representation of the goal observation \\

$\rho(h)$
& Normalized Lorentz geodesic radius used as a diagnostic
& $\log^c_x(y)$
& Lorentz logarithmic map from point $x$ to point $y$ \\

$\widehat{h}_{t+k}$
& Future latent state predicted by the latent dynamics
& $\delta_p^g$
& Normalized parent-to-goal distance used for adaptive aperture \\

$g$
& Goal observation or goal state
& $\rho(s,g)$
& Task progress score of state $s$ with respect to goal $g$ \\

$s_i \prec_g s_j$
& Goal-conditioned progress relation indicating that $s_j$ has higher progress than $s_i$
& $(h_p,h_c,h_g)$
& Parent, child, and goal latents constructed from hindsight trajectories \\

$\Delta$
& Temporal offset used when constructing parent-child state pairs
& $\theta_{c,g}^{(t)}$
& Angular deviation between parent-child and parent-goal directions \\

$A(\delta_p^g)$
& entailment cone aperture determined by parent-to-goal distance
& $m_g$
& goal-progress margin within the entailment cone \\

$\preceq_{\mathrm{cone}}$
& directed progress relation induced by entailment cones in the hyperbolic latent space
& $\mathcal{L}_{\mathrm{cone}}$
& entailment cone constraint loss induced by progress \\

$\mathcal{L}_{\mathrm{ctr}}$
& Hyperbolic contrastive loss in the Lorentz model
& $L$
& planning horizon at inference time \\

$d^g_L$
& hyperbolic distance between the rollout endpoint and the goal
& $C_{\mathrm{best}}$
& cost of the intermediate state closest to the goal among candidate trajectories \\

$C_{\mathrm{mean}}$
& average goal-distance cost over an entire candidate trajectory
& $\beta_T,\beta_b,\beta_m$
& planning-cost weights for terminal, best, and mean trajectory costs \\

$\mathrm{SR}$
& Success Rate, denoting the success rate on test tasks
& $\mathrm{MSE}$
& Mean Squared Error, used to evaluate the readability of physical state information from the latent representation \\

\bottomrule
\end{tabular}
\end{table*}

\subsection{Ablation Table Abbreviations}
\label{app:ablation_abbreviations}

Table~\ref{tab:ablation_abbreviations} summarizes the compact labels used in the main ablation tables.

\begin{table*}[t]
\centering
\caption{Abbreviations used in the main ablation tables.}
\label{tab:ablation_abbreviations}
\scriptsize
\setlength{\tabcolsep}{5pt}
\renewcommand{\arraystretch}{1.05}
\begin{tabular}{lp{0.31\textwidth}lp{0.31\textwidth}}
\toprule
\textbf{Label} & \textbf{Meaning} & \textbf{Label} & \textbf{Meaning} \\
\midrule
w/o & corresponding component removed
& Hyp. & Lorentz hyperbolic latent space \\
Pred. & $\mathcal{L}_{\mathrm{pred}}$
& Euc. & Euclidean latent-space replacement \\
Ctr. & $\mathcal{L}_{\mathrm{ctr}}$
& Rev. & reversed progress order \\
Cone. & $\mathcal{L}_{\mathrm{cone}}$
& Rand. & random state-pair sampling \\
Reg. & $\mathcal{L}_{\mathrm{sig}}$
& Fixed & fixed cone aperture \\
Correct Order & temporally correct parent-child order
& Same Traj. Pair & pair sampled from the same trajectory \\
Adaptive Cone & adaptive cone aperture
& Dist. & goal-distance progress term \\
Angle / Ang. & angular cone term
& Seg. Cone & cone variant additionally using $h_{t+1}$ \\
$D_L^g$ & terminal goal-distance cost
& $C_{\mathrm{best}}$ / $C_{\mathrm{mean}}$ & best-intermediate / mean goal-distance cost \\
\bottomrule
\end{tabular}
\end{table*}

\section{Theoretical Discussion}
\label{app:theoretical_discussion}

This section supplements the formal discussion of goal-conditioned progress order presented in the main text,
and explains how the entailment cone constraints in ProWorld realize a directional progress relation in the hyperbolic latent space.

\subsection{Asymmetry of Goal-Conditioned Progress}
\label{app:progress_asymmetry}

Let $\mathcal{S}$ denote the state space and $\mathcal{G}$ denote the goal space.
For a given goal $g \in \mathcal{G}$, the goal-conditioned progress potential function is defined as
\[
\rho_g : \mathcal{S} \rightarrow \mathbb{R}
\]
where a larger $\rho_g(s)$ indicates that state $s$ represents higher task progress toward achieving goal $g$.

Given a goal-conditioned hindsight trajectory segment associated with goal $g$,
\[
\tau = (s_0, s_1, \ldots, s_T; g),
\]
we write $s_i \prec_g s_j$ for two states $s_i$ and $s_j$ on the same trajectory if there exists $\epsilon > 0$ such that
\[
i < j
\quad \text{and} \quad
\rho_g(s_j) \geq \rho_g(s_i) + \epsilon.
\]
This relation means that $s_j$ has higher goal-conditioned progress than $s_i$.

\paragraph{Proof of asymmetry.}
If $s_i \prec_g s_j$ holds, then by definition we have
\[
i < j
\]
and
\[
\rho_g(s_j) \geq \rho_g(s_i) + \epsilon
\]
If the reverse relation $s_j \prec_g s_i$ also holds, then the following conditions must also hold:
\[
j < i
\]
and
\[
\rho_g(s_i) \geq \rho_g(s_j) + \epsilon
\]
However, $i < j$ and $j < i$ cannot hold simultaneously.
Meanwhile, adding the two progress inequalities yields
\[
\rho_g(s_j) + \rho_g(s_i)
\geq
\rho_g(s_i) + \rho_g(s_j) + 2\epsilon
\]
that is
\[
0 \geq 2\epsilon
\]
This contradicts $\epsilon > 0$.
Therefore, $s_i \prec_g s_j$ and $s_j \prec_g s_i$ cannot hold simultaneously,
and the goal-conditioned progress relation is asymmetric.

\subsection{Cone Realization in Hyperbolic Latent Space}
\label{app:cone_realization}

ProWorld realizes the above directional progress relation with a goal-anchored entailment cone. For a parent-child-goal triple $(h^p,h^c,h_g)$ from a goal-conditioned hindsight trajectory segment, where $h^p$ is earlier than $h^c$, we compute two tangent directions at the parent:
\[
v_c=\log^c_{h^p}(h^c),
\qquad
v_g=\log^c_{h^p}(h_g).
\]
The child is encouraged to move along the parent-to-goal direction through the angular constraint
\[
\theta_{c,g}
\leq
A(\delta_p^g),
\]
where $\theta_{c,g}$ is the angle between $v_c$ and $v_g$, and $\delta_p^g$ is the normalized parent-to-goal distance used by the adaptive aperture. Progress is further constrained by a goal-distance margin:
\[
d_{\mathbb{H}}(h^c,h_g)+m_g
\leq
d_{\mathbb{H}}(h^p,h_g).
\]
We define
\[
h^p \preceq_{\mathrm{cone}}^g h^c
\]
when both constraints hold.

This relation is explicitly goal-conditioned because both the angular direction and the distance margin depend on $h_g$. The same parent-child pair can therefore receive different supervision under different hindsight goals. Its asymmetry follows from the goal-distance margin: if $h^p \preceq_{\mathrm{cone}}^g h^c$ holds, then $h^c$ is at least $m_g$ closer to $h_g$ than $h^p$. The reverse relation $h^c \preceq_{\mathrm{cone}}^g h^p$ would require $h^p$ to be at least $m_g$ closer to the same goal than $h^c$, which is impossible for $m_g>0$. Thus, the cone constraint is not an ordinary symmetric distance regularizer; it imposes a directed, goal-dependent progress relation in the hyperbolic latent space.

\section{Environment \& Datasets}
\subsection{Details of Compared Methods}

This section briefly introduces the baseline methods used in Tab.~\ref{tab:main_results}.
Overall, these methods fall into three categories:
latent world model-based visual planning methods,
goal-conditioned offline reinforcement learning methods,
and simple random-action baselines.

\paragraph{LeWM.}
LeWM~\cite{lewm} is a latent world model method based on a joint-embedding predictive architecture.
It learns a visual encoder and action-conditioned latent dynamics from image-action trajectories,
and performs goal-conditioned planning at inference time via latent rollout over candidate action sequences.
Specifically, LeWM maps the current observation and goal image into the latent space,
and selects the action sequence whose predicted terminal latent state is closest to the goal latent representation.

\paragraph{C-JEPA.}
C-JEPA~\cite{c-jepa} is a JEPA-style visual representation learning baseline
that learns discriminative latent representations of future states via contrastive or predictive objectives.
In our comparison, C-JEPA is used to examine whether generic joint-embedding representations suffice for goal-conditioned visual planning.

% \paragraph{GCBC.}
% \textcolor{red}{From here through the GCIQL description below, revisions are needed: the methods do not operate in state space—refer to LeWM}
% Goal-Conditioned Behavioral Cloning (GCBC)~\cite{ghosh2021gcsl} is the most straightforward goal-conditioned imitation learning baseline.
% The method takes the current observation and goal as input, and fits expert actions in offline data via supervised learning.
% GCBC learns no explicit dynamics model and performs no candidate trajectory search at inference time,
% so its performance depends primarily on whether the offline data contains local behavior patterns that match the test goals.

\paragraph{GCIVL.}
Goal-Conditioned Implicit V-Learning (GCIVL)~\cite{ogbench} is one of the goal-conditioned offline RL baselines provided in OGBench.
It learns a value function under the goal-conditioned setting and performs policy learning using state transitions and goal-relabeling signals from offline data.
Compared to GCBC, GCIVL can leverage value function estimates to distinguish the long-term value of different actions or states for goal completion,
yet its core remains policy learning rather than explicit planning through a visual latent world model.

\paragraph{GCIQL.}
Goal-Conditioned Implicit Q-Learning (GCIQL)~\cite{ogbench,kostrikov2022iql} is an offline RL baseline obtained by adapting implicit Q-learning to the goal-conditioned setting in OGBench.
It learns a goal-conditioned Q-function, value function, and policy,
and mitigates out-of-distribution action estimation through a conservative offline value learning mechanism.

\paragraph{PLDM.}
PLDM~\cite{PLDM} is a planning-oriented latent dynamics baseline for offline reward-free reinforcement learning.
It learns a compact latent transition model from offline trajectories and uses the learned dynamics to evaluate candidate action sequences.
In our comparison, PLDM serves as a representative latent-dynamics planning method that does not explicitly impose goal-conditioned hyperbolic progress structure.

\paragraph{EB-JEPA.}
EB-JEPA~\cite{EB-JEPA} is a representation learning and world model framework based on an energy-based joint-embedding predictive architecture.
It predicts future representations in representation space,
thereby avoiding the high-dimensional generative burden of reconstructing future frames directly in pixel space.
In our experiments, EB-JEPA serves as the JEPA-type world model baseline,
used to compare general representation-predictive world models against the hyperbolic progress modeling in ProWorld.

\paragraph{TD-MPC2.}
TD-MPC2~\cite{TD-MPC2} is a model-based reinforcement learning method designed for continuous control tasks.
It learns a compact latent world model and combines model predictive control to optimize action sequences in latent space.
Since TD-MPC2 is primarily designed for reward-driven continuous control,
whereas the tasks in this paper emphasize image goal-conditioned planning,
this baseline is used to examine the adaptability of general latent MPC methods to visual goal planning scenarios.

\paragraph{Sub-JEPA.}
Sub-JEPA~\cite{Sub-JEPA} extends JEPA-style world modeling with subspace Gaussian regularization to improve representation stability during end-to-end dynamics learning.
It preserves the latent prediction paradigm while regularizing feature subspaces against collapse or degenerate covariance structure.
We include Sub-JEPA to compare ProWorld with a stabilization-focused JEPA baseline that does not explicitly encode goal-conditioned progress order.

\paragraph{GeoWorld-Style.}
GeoWorld~\cite{zhang2026geoworld} introduces hyperbolic world representations on top of V-JEPA 2.
Its core idea is to first preserve JEPA-style future prediction through teacher-forcing and rollout representation supervision, and then use Geometric Reinforcement Learning (GRL) to constrain rollout geometry in hyperbolic space, including hyperbolic energy minimization and triangle-inequality regularization.
Because the official implementation, pretrained weights, and complete evaluation scripts of the original work were not publicly available at the time of submission, and because its original experimental setup differs from our offline visual goal-planning tasks, we do not report official GeoWorld results.
To provide a nearest-neighbor comparison, we implement a GeoWorld-Style proxy baseline.
This baseline preserves three main design ideas from GeoWorld: mapping JEPA latent states into hyperbolic space, using teacher-forcing/rollout representation loss to maintain one-step and multi-step prediction consistency, and adding GRL-style hyperbolic geometric regularization.
The key difference is that we initialize the visual encoder and latent dynamics from official LeWM pretrained weights, rather than using V-JEPA 2 pretrained weights; we then finetune this baseline under the same training data, image preprocessing, action-conditioned rollout, and CEM planning budget as in our experiments.

\begin{GeoCodeBlock}
def GeoWorldStyle(obs, act, goal, lambd=0.02):
    """obs: visual sequence; act: action sequence."""
    lewm = load_official_lewm_checkpoint()
    z = lewm.encoder(obs)
    z_hat = lewm.predictor(z, act)

    # teacher-forcing / rollout loss
    tfr = mse(z_hat[:, :-1], z[:, 1:])

    # GeoWorld-style hyperbolic GRL
    h = lorentz_project(z)
    h_hat = lorentz_project(z_hat)
    h_goal = lorentz_project(lewm.encoder(goal))
    grl = hyperbolic_energy(h_hat, h[:, 1:])
    grl += triangle_reg(h, h_hat, h_goal)

    return tfr + lambd * grl
\end{GeoCodeBlock}

\paragraph{Random.}
Random denotes the random action baseline.
This method uses no training data, goal information, or learned model,
and instead samples actions uniformly at random from the action space at each timestep.
This result serves as a lower-bound reference for task difficulty.

\subsection{Datasets}
\label{sec:dataset}

\paragraph{Training Goal Construction.}
During training, ProWorld does not require explicit success labels for every offline trajectory.
For goal-conditioned supervision, we construct hindsight goal contexts from future states within the same trajectory.
Given a trajectory segment, a future observation is treated as the visual goal, and temporally ordered state pairs before that future state are used as weak progress-order supervision.
Thus, the progress-pair construction uses a goal-conditioned hindsight segment in which the later state is reachable within the same recorded trajectory, rather than an externally labeled episode-level success.

This distinction is particularly important for Cube-S play data.
The Cube-S dataset is not collected toward a single fixed goal; instead, it contains play-style manipulation behaviors.
For training, the goal context is obtained from future frames in the same trajectory, including the corresponding cube pose information when available, and progress pairs are sampled according to temporal order relative to that hindsight goal.
Evaluation success, by contrast, is determined by the environment termination signal after writing the dataset goal pose into the environment, as described in the evaluation protocol below.

\paragraph{PushT.}
PushT~\cite{Dino-wm} is a continuous 2D planar manipulation task in which the agent must push and rotate a T-shaped object to a target pose through contact interactions. The task exhibits pronounced contact dynamics and long-horizon error accumulation, making it a standard benchmark for evaluating a world model's ability to capture object interactions, motion trends, and multi-step planning. We adopt the same data setting as Zhou et al.~\cite{Dino-wm}, comprising 20,000 expert trajectories with an average length of 196 steps.

\paragraph{OGBench-Cube-Single-Play (Cube-S).}
OGBench-Cube-Single-Play~\cite{ogbench} is a robotic manipulation task in OGBench that evaluates an agent's basic object manipulation capability. The environment is built on a UR5e robotic arm in MuJoCo, requiring the arm to perform pick-and-place operations via end-effector control. We adopt the \texttt{visual-cube-single-play-v0} configuration, where the maximum episode length for evaluation is 200 steps, and the official offline dataset contains 1M transitions across 1,000 episodes with a data episode length of 1,000. The dataset is collected by a non-Markovian scripted policy with temporally correlated noise, constituting play-style data. Since the trajectories are not generated toward a single fixed goal but instead cover a variety of composable local manipulation behaviors, this task tests whether a world model can learn local object manipulation dynamics and compositional structure from offline visual trajectories.

\paragraph{OGBench-Scene-Play (Scene).}
OGBench-Scene-Play~\cite{ogbench} is a compositional visual manipulation task involving multiple object types, including cubes, buttons, drawers, and windows. We adopt the \texttt{visual-scene-play-v0} dataset and evaluate in the corresponding \texttt{visual-scene-v0} environment. The official offline dataset contains 1M transitions across 1,000 episodes with a data episode length of 1,000 transitions. Unlike purely continuous progress tasks, Scene combines continuous object displacement or articulation with discrete button states, and the final success signal remains a sparse environment termination event. We use a hindsight goal offset of 50 steps and an evaluation budget of 100 interaction steps. This task therefore evaluates whether the learned visual world model can support mixed discrete-continuous, multi-object goal reaching under a longer compositional horizon.

\paragraph{OGBench-AntMaze-Large-Navigate (AntMaze-L).}
OGBench-AntMaze-Large-Navigate~\cite{ogbench} is a long-horizon continuous control navigation task in OGBench, requiring the agent to control an 8-DoF Ant robot to navigate through a maze and reach a specified goal location. The task combines low-level locomotion control with high-level spatial navigation structure, making it considerably more challenging than simple 2D point navigation. We adopt the \texttt{antmaze-large-navigate-v0} configuration, where the maximum episode length for evaluation is 1,000 steps, and the official offline dataset contains 1M transitions across 1,000 episodes with a data episode length of 1,000. The Navigate dataset is collected by a noisy expert navigation policy, with trajectories primarily covering locomotion from diverse start positions to diverse goal locations. This task is well suited for evaluating a world model's capabilities in long-horizon prediction, continuous control dynamics modeling, and cross-region navigation structure learning.

\begin{table*}[t]
\centering
\caption{Sensitivity analysis of temporal offset $\Delta$. Bold numbers indicate the best Success Rate (SR).}
\label{tab:sensitivity_delta}
\scriptsize
\setlength{\tabcolsep}{8pt}
\begin{tabular}{lccccc}
\toprule
$\Delta$ & \textbf{1} & \textbf{2} & \textbf{4} & \textbf{8} & \textbf{10} \\
\midrule
SR (\%) & 70 & \textbf{78} & 76 & 68 & 62 \\
\bottomrule
\end{tabular}
\end{table*}

\begin{table*}[t]
\centering
\caption{Sensitivity analysis of curvature $c$. Bold numbers indicate the best Success Rate (SR).}
\label{tab:sensitivity_curvature}
\scriptsize
\setlength{\tabcolsep}{8pt}
\begin{tabular}{lccccc}
\toprule
$c$ & \textbf{0.1} & \textbf{0.5} & \textbf{1.0} & \textbf{2.0} & \textbf{Learnable} \\
\midrule
SR (\%) & 68 & 74 & \textbf{78} & 72 & 76 \\
\bottomrule
\end{tabular}
\end{table*}

\begin{table*}[t]
\centering
\caption{Sensitivity analysis of planning horizon $L$. Bold numbers indicate the best Success Rate (SR).}
\label{tab:sensitivity_planning_horizon}
\scriptsize
\setlength{\tabcolsep}{8pt}
\begin{tabular}{lccccc}
\toprule
$L$ & \textbf{3} & \textbf{5} & \textbf{10} & \textbf{15} & \textbf{20} \\
\midrule
SR (\%) & 66 & \textbf{78} & 74 & 68 & 60 \\
\bottomrule
\end{tabular}
\end{table*}

\section{Additional Ablation Experiments}
\label{sec:additional_ablation_experiments}
\begin{figure}[t]
    \centering

    \begin{subfigure}[t]{0.48\linewidth}
        \centering
        \includegraphics[width=\linewidth]{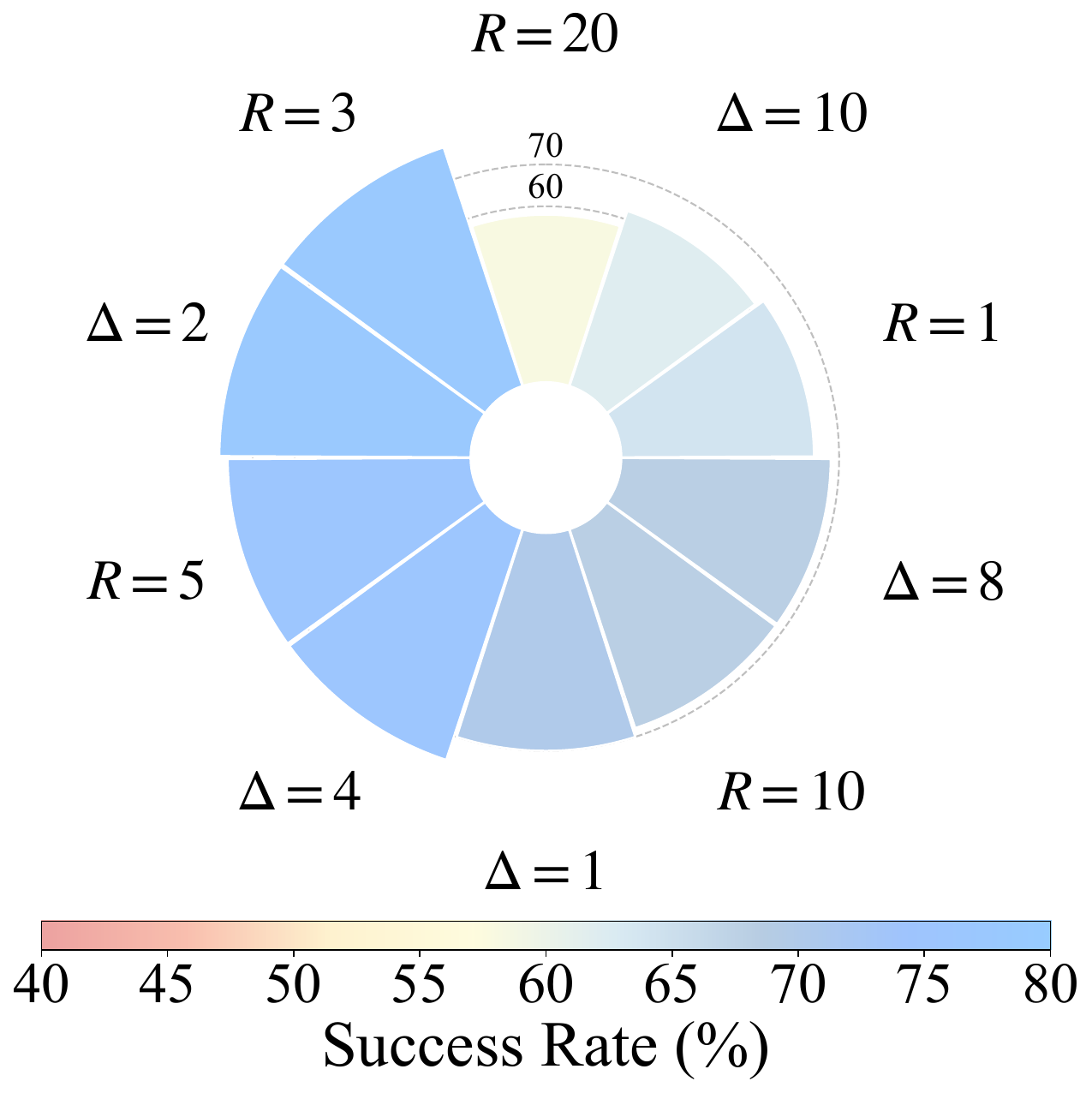}
        \caption{Sensitivity to temporal offset $\Delta$.}
    \end{subfigure}
    \hfill
    \begin{subfigure}[t]{0.48\linewidth}
        \centering
        \includegraphics[width=\linewidth]{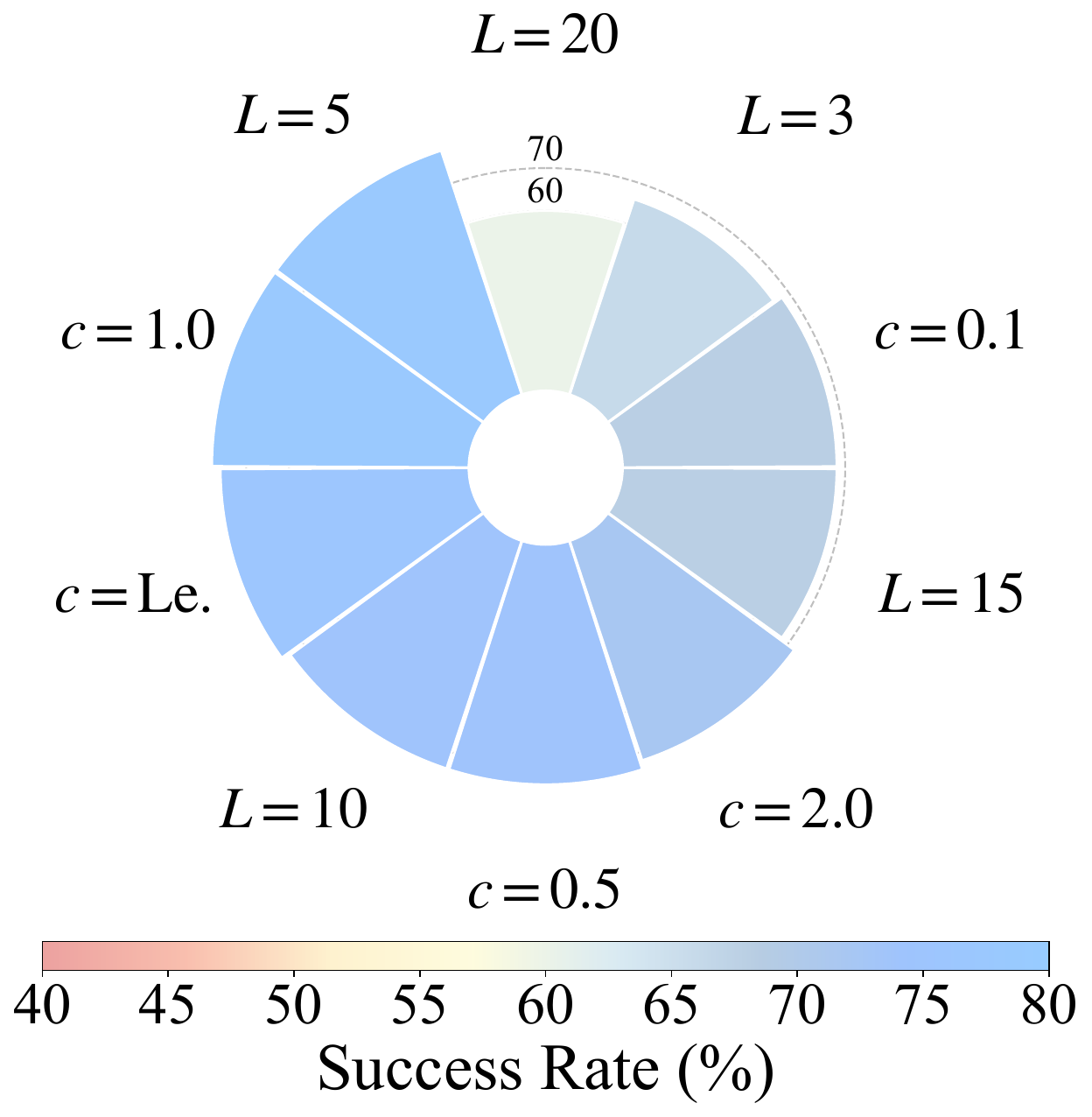}
        \caption{Sensitivity to curvature $c$ and planning horizon $L$.}
    \end{subfigure}

    \vspace{2mm}

    \begin{subfigure}[t]{0.62\linewidth}
        \centering
        \includegraphics[width=\linewidth]{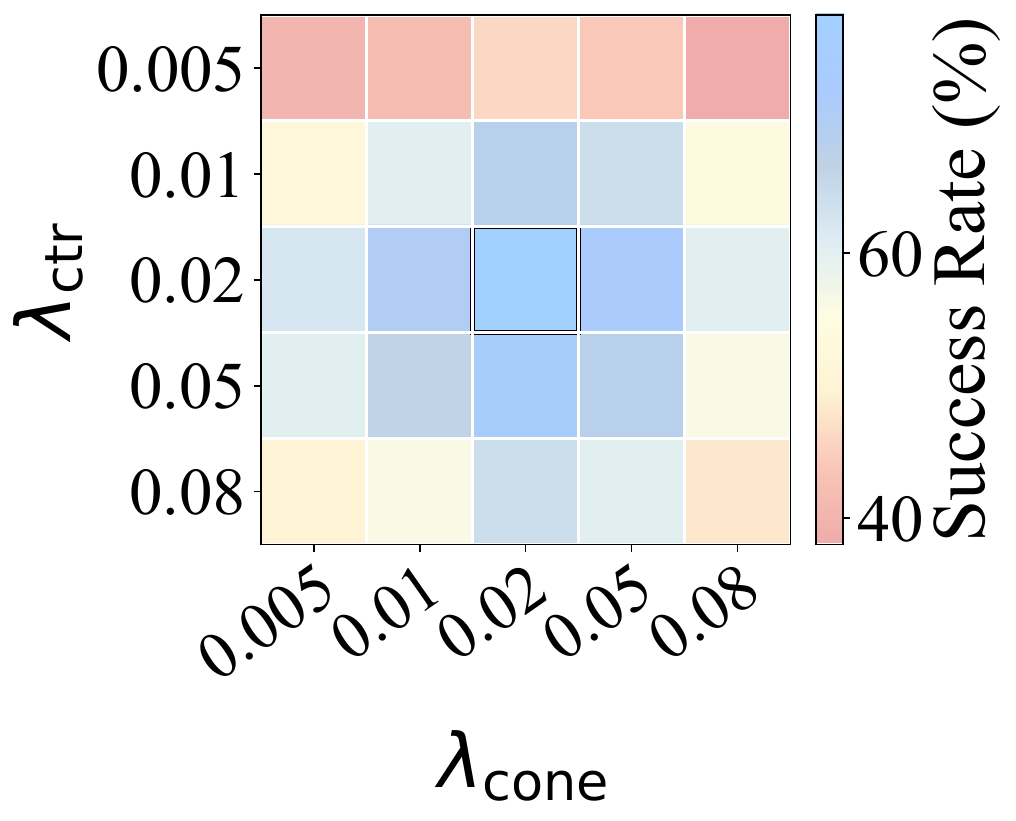}
        \caption{Two-dimensional joint sensitivity of $\lambda_{\mathrm{ctr}}$ and $\lambda_{\mathrm{cone}}$.}
    \end{subfigure}

    \caption{Visualization of supplementary hyperparameter sensitivity analyses. The polar bar charts show the effect of temporal-offset, curvature, and planning-horizon variations on SR, where both color and radial length represent SR; the heatmap shows the joint effect of $\lambda_{\mathrm{ctr}}$ and $\lambda_{\mathrm{cone}}$, with colors closer to high-value regions indicating higher success rates.}
    \label{fig:supp_sensitivity_visualization}
\end{figure}

\begin{figure}[t]
    \centering
    \begin{minipage}{0.48\linewidth}
        \centering
        \includegraphics[width=\linewidth]{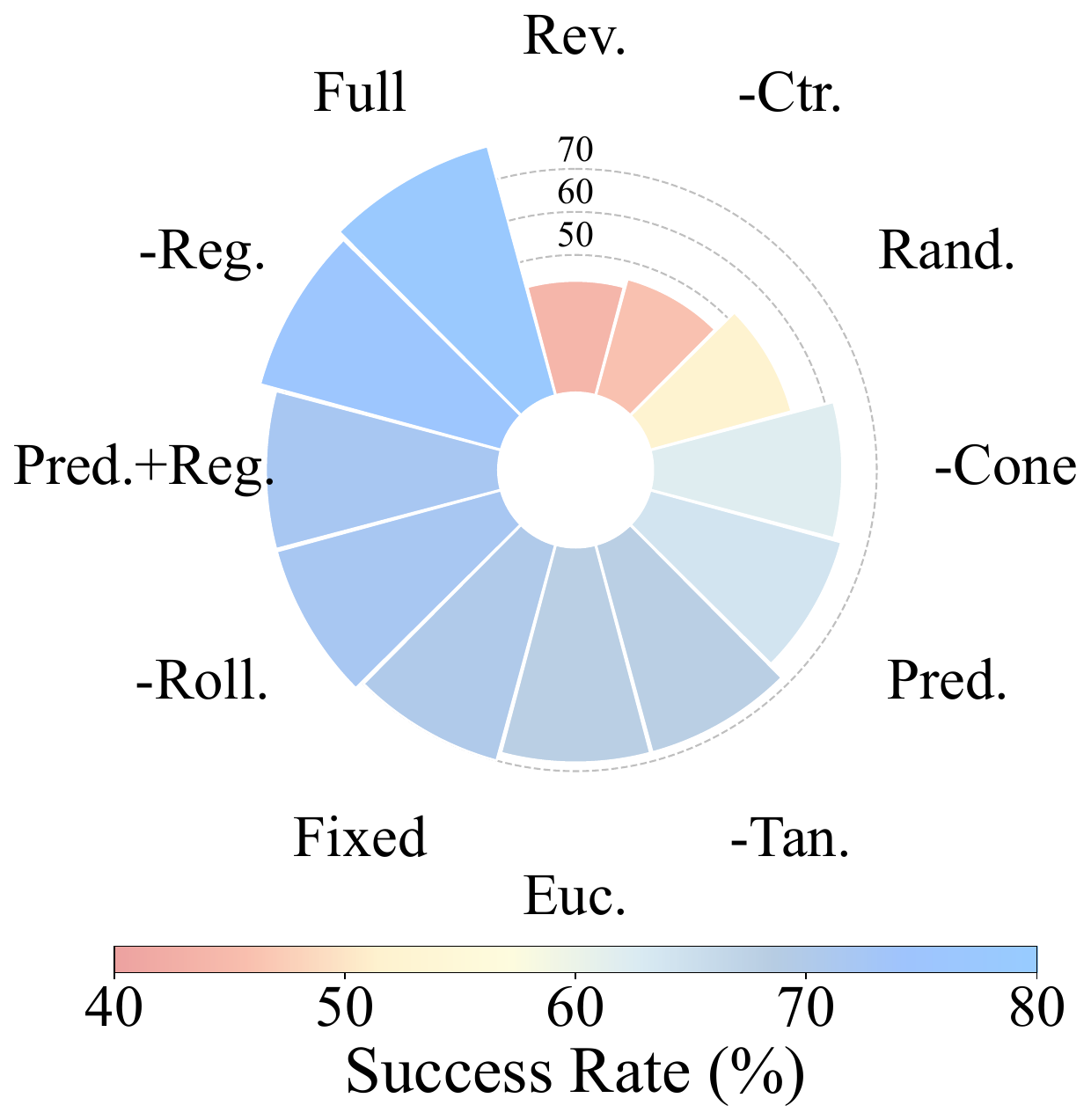}
        \centerline{(a)}
    \end{minipage}
    \hfill
    \begin{minipage}{0.48\linewidth}
        \centering
        \includegraphics[width=\linewidth]{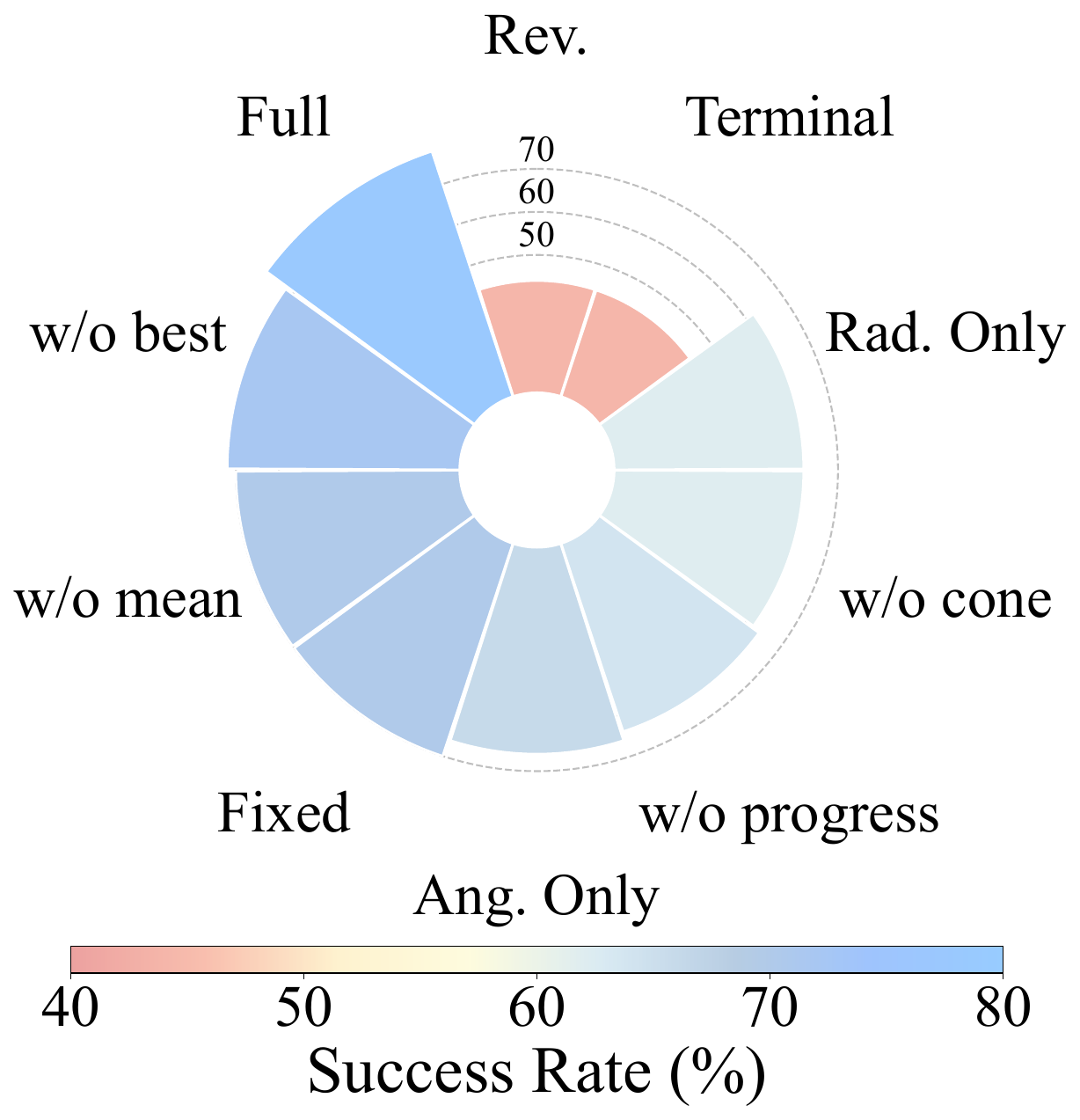}
        \centerline{(b)}
    \end{minipage}
    \caption{Polar bar visualization of the ablation studies in the main text. Both color and radial length represent planning SR, with the color scale ranging from red (low) to blue (high). (a) Summarizes the learning objective ablation and the geometry-and-progress-order ablation; (b) summarizes the entailment cone design ablation and the planning objective ablation. All results are obtained on Cube-S; Full denotes the complete ProWorld.}
    \label{fig:main_ablation_polar}
\end{figure}

This section further analyzes the effects of key hyperparameters and training objective weights on planning performance in ProWorld.
All experiments are conducted on Cube-S using the same training data, network architecture, and planning budget as in the main text. In addition to the tabular results, Fig.~\ref{fig:supp_sensitivity_visualization} and Fig.~\ref{fig:main_ablation_polar}
further summarize the main ablation and sensitivity results in polar chart form,
to more intuitively illustrate the effect of different modules and hyperparameter settings on SR.

\subsection{Sensitivity to Temporal Offset $\Delta$}

The entailment cone constraint constructs parent-child-goal triples $(h_t, h_{t+\Delta}, h_g)$ from the same goal-conditioned hindsight trajectory segment, where $\Delta$ controls the temporal span of the progress supervision.
As shown in Tab.~\ref{tab:sensitivity_delta}, $\Delta=2$ achieves the best SR of $78\%$; $\Delta=4$ yields $76\%$, which is close to the best.
In contrast, $\Delta=1$ drops to $70\%$, and further increasing to $\Delta=8$ and $\Delta=10$ reduces SR to $68\%$ and $62\%$, respectively.

This result indicates that progress supervision requires a moderate temporal interval.
When $\Delta$ is too small, the semantic progress between adjacent states is weak, and parent-child pairs fail to provide clear hierarchical structure signals;
when $\Delta$ is too large, local reachability between state pairs diminishes, the cone constraints may become overly sparse, and the quality of progress modeling in the latent space degrades.
Therefore, $\Delta=2$ strikes a favorable balance between local dynamics consistency and goal-conditioned progress supervision.

\subsection{Sensitivity to Curvature $c$}

Tab.~\ref{tab:sensitivity_curvature} compares the effects of different hyperbolic curvature settings.
At $c=1.0$, the model achieves the highest SR of $78\%$;
learnable curvature achieves $76\%$, close to the fixed optimal curvature.
Smaller curvatures $c=0.1$ and $c=0.5$ yield $68\%$ and $74\%$, respectively, while larger curvature $c=2.0$ yields $72\%$.

These results indicate that the scale of the negatively curved latent space affects the representational organization of ProWorld.
When the curvature is too small, the hierarchical expressiveness of the hyperbolic space weakens, making it difficult for the model to adequately capture the goal-conditioned progress structure;
when the curvature is too large, distances and angular variations may become excessively sharp, disrupting the coordination between dynamics prediction and cone constraints.
Meanwhile, the learnable curvature performs comparably to $c=1.0$, suggesting that the fixed curvature used in this work already provides effective geometric inductive bias without relying heavily on an additional curvature learning mechanism.

\subsection{Joint Sensitivity to $\lambda_{\mathrm{ctr}}$ and $\lambda_{\mathrm{cone}}$}

$\lambda_{\mathrm{ctr}}$ and $\lambda_{\mathrm{cone}}$ control the weights of the hyperbolic contrastive loss and the entailment cone loss, respectively.
Both jointly influence goal discriminability and progress structure in the latent space, so we conduct a two-dimensional sensitivity analysis over them.
As shown in Tab.~\ref{tab:sensitivity_ctr_cone_matrix}, when both take moderate values, i.e., $\lambda_{\mathrm{ctr}}=0.03$ and $\lambda_{\mathrm{cone}}=0.02$, the model achieves the highest SR of $78\%$.

Overall, weights that are either too small or too large lead to performance degradation.
When $\lambda_{\mathrm{ctr}}=0.00$, the SR remains low across all $\lambda_{\mathrm{cone}}$ values, indicating that removing contrastive discrimination severely impairs the separability of future states in the latent space.
As $\lambda_{\mathrm{ctr}}$ increases to $0.01$ and $0.03$, performance improves markedly, suggesting that moderate contrastive regularization helps form more plannable goal-conditioned representations.
However, as $\lambda_{\mathrm{ctr}}$ further increases to $0.05$ or $0.10$, performance degrades again, indicating that overly strong contrastive constraints may suppress the dynamics prediction objective.

Similarly, with $\lambda_{\mathrm{ctr}}=0.03$ fixed, increasing $\lambda_{\mathrm{cone}}$ from $0.005$ to $0.02$ raises SR from $70\%$ to $78\%$;
but further increasing it to $0.05$ and $0.10$ drops SR to $72\%$ and $60\%$, respectively.
This indicates that the entailment cone constraint effectively introduces progress order information, but an overly strong cone constraint leads to an excessively rigid latent space geometry.
Thus, the optimal performance of ProWorld stems from a balance between contrastive discriminability and progress entailment structure, rather than simply increasing either weight alone.

\begin{table*}[t]
\centering
\caption{Two-dimensional sensitivity analysis of $\lambda_{\mathrm{ctr}}$ and $\lambda_{\mathrm{cone}}$. Each entry reports the Success Rate (SR \%). Bold number indicates the best result.}
\label{tab:sensitivity_ctr_cone_matrix}
\scriptsize
\setlength{\tabcolsep}{8pt}
\begin{tabular}{cccccc}
\toprule
\multirow{2}{*}{$\lambda_{\mathrm{ctr}}$}
& \multicolumn{5}{c}{$\lambda_{\mathrm{cone}}$} \\
\cmidrule(lr){2-6}
& \textbf{0.00} & \textbf{0.005} & \textbf{0.02} & \textbf{0.05} & \textbf{0.10} \\
\midrule
0.00 & 40 & 42 & 46 & 44 & 38 \\
0.01 & 52 & 60 & 68 & 64 & 54 \\
0.03 & 62 & 70 & \textbf{78} & 72 & 60 \\
0.05 & 60 & 66 & 74 & 68 & 56 \\
0.10 & 50 & 56 & 64 & 60 & 48 \\
\bottomrule
\end{tabular}
\end{table*}

\begin{table*}[t]
\centering

\begin{minipage}[t]{0.48\textwidth}
\centering
\caption{Training hyperparameters of ProWorld.}
\label{tab:training_hparams}
\small
\setlength{\tabcolsep}{5pt}
\resizebox{\linewidth}{!}{
\begin{tabular}{lcc}
\toprule
\textbf{Term} & \textbf{Symbol} & \textbf{Value} \\
\midrule
Hyperbolic prediction & $\lambda_{\mathrm{pred}}$ & 1.0 \\
Lorentz contrastive learning & $\lambda_{\mathrm{ctr}}$ & 0.03 \\
Entailment cone regularization & $\lambda_{\mathrm{cone}}$ & 0.02 \\
SIGReg regularization & $\lambda_{\mathrm{sig}}$ & 0.06 \\
\midrule
Minimum aperture & $A_{\min}$ & 0.10 \\
Maximum aperture & $A_{\max}$ & 0.95 \\
Angle penalty weight & $w_{\theta}$ & 1.0 \\
Goal-distance penalty weight & $w_g$ & 1.0 \\
Goal-progress margin & $m_g$ & 0.03 \\
Temporal offset & $\Delta$ & 2 \\
\bottomrule
\end{tabular}
}
\end{minipage}
\hfill
\begin{minipage}[t]{0.48\textwidth}
\centering
\caption{Planning hyperparameters of ProWorld.}
\label{tab:planning_hparams}
\scriptsize
\setlength{\tabcolsep}{2.5pt}
\resizebox{\linewidth}{!}{
\begin{tabular}{lcccc}
\toprule
\textbf{Parameter} & \textbf{PushT} & \textbf{Cube-S} & \textbf{AntMaze-L} & \textbf{Scene} \\
\midrule
Planner & CEM & CEM & CEM & CEM \\
Candidate action sequences & 300 & 300 & 300 & 300 \\
Planning iterations & 30 & 30 & 30 & 30 \\
Elite samples $k$ & 30 & 30 & 30 & 30 \\
Variance scale & 1.0 & 1.0 & 1.0 & 1.0 \\
Planning horizon $L$ & 5 & 5 & 15 & 5 \\
Receding horizon & 5 & 5 & 5 & 5 \\
Action block size & 5 & 5 & 1 & 5 \\
Evaluation budget & 50 & 50 & 100 & 100 \\
\midrule
Terminal cost $\beta_T$ & 1.0 & 1.0 & 1.0 & 1.0 \\
Best intermediate cost $\beta_b$ & 0.35 & 0.35 & 0.35 & 0.35 \\
Mean rollout cost $\beta_m$ & 0.15 & 0.15 & 0.15 & 0.15 \\
Progress margin $\delta$ & 0.0 & 0.0 & 0.0 & 0.0 \\
\bottomrule
\end{tabular}
}
\end{minipage}

\end{table*}

\subsection{Sensitivity to Planning Horizon $L$}

Tab.~\ref{tab:sensitivity_planning_horizon} analyzes the effect of the planning horizon $L$ during inference.
At $L=5$, the model achieves the best SR of $78\%$.
A shorter planning horizon of $L=3$ yields only $66\%$, indicating that when candidate action sequences are too short, the planner cannot sufficiently leverage goal-conditioned progress information.
When the planning horizon is extended to $L=10$, SR remains at $74\%$, suggesting that ProWorld maintains good multi-step prediction stability under moderate horizon extension.
However, as $L$ increases further to $15$ and $20$, SR drops to $68\%$ and $60\%$, respectively.

This result indicates that a longer planning horizon does not necessarily improve planning performance.
On one hand, an overly short horizon limits the planner's ability to assess long-range goal progress;
on the other hand, an overly long horizon increases the accumulated error of the latent rollout, causing candidate trajectory scores to be affected by prediction drift.
Therefore, $L=5$ provides a favorable planning length under the current task and model configuration, enabling the model to exploit intermediate progress signals while avoiding the error accumulation introduced by excessively long rollouts.

% preamble:
% \usepackage{graphicx}
% \usepackage{subcaption}

\section{Analysis of Progress Structure}
\label{app:progress_structure_analysis}

\subsection{Evaluation Metrics}
\label{app:evaluation_metrics}
\paragraph{SR.}
Success Rate (SR) measures the final completion rate of a method on goal-conditioned visual planning tasks.
For each test episode, we provide the initial observation $o^{(n)}_0$ and the goal observation $g^{(n)}$,
the model executes actions according to learned latent dynamics or a policy and determines within a fixed interaction budget whether the goal has been reached.
Let the test set contain $N$ episodes, where $\tau^{(n)}$ denotes the execution trajectory of the $n$-th episode,
$\mathrm{Succ}(\tau^{(n)}, g^{(n)})\in\{0,1\}$ indicates whether the trajectory satisfies the success condition defined by the environment; SR is then defined as
\begin{equation}
    \mathrm{SR}
    =
    \frac{1}{N}
    \sum_{n=1}^{N}
    \mathbb{I}
    \left[
    \mathrm{Succ}(\tau^{(n)}, g^{(n)}) = 1
    \right]
    \times 100\%
\end{equation}
where $\mathbb{I}[\cdot]$ denotes the indicator function. Thus, a higher SR reflects a stronger ability of the model to achieve goals within a limited number of planning steps.

\paragraph{Success Criterion and Goal Binding.}
\label{app:success_goal_binding}
It is important to note that the SR reported in this paper is not computed from final-frame visual similarity, latent distance, or an additional hand-written task threshold.
During interactive evaluation, we read the native \texttt{terminated} signal returned by the environment at each step within a fixed evaluation budget, and count an episode as successful if \texttt{terminated=True} occurs at any step.
In other words, once a trajectory reaches the success condition internally defined by the environment at some point within the budget, the episode is accumulated as a success even if the state subsequently deviates from the goal.
Formally, let $e_t^{(n)}\in\{0,1\}$ denote the \texttt{terminated} signal returned by the environment at interaction step $t$ of the $n$-th episode. Then
\begin{equation}
    \mathrm{Succ}(\tau^{(n)}, g^{(n)})
    =
    \mathbb{I}
    \left[
    \exists\, t \le B
    \ \text{s.t.}\ 
    e_t^{(n)} = 1
    \right]
\end{equation}
where $B$ denotes the evaluation budget.

For OGBench tasks, the evaluation script randomly samples a valid start row and uses a future state from the same episode as the hindsight goal. The environment is initialized from the dataset state at the start row, while the goal image and goal state are written into the evaluation world information used by the planner and model. At every interaction step, success is read from \texttt{world.terminateds}; an episode is counted as successful if any environment termination signal becomes true within the evaluation budget.

For Cube-S, the script restores the initial state using \texttt{qpos} and \texttt{qvel}, and binds the goal by writing the future cube target position and orientation from \texttt{goal\_privileged\_block\_0\_pos} and \texttt{goal\_privileged\_block\_0\_quat}. We use a goal offset of 25 steps and an evaluation budget of 50 steps. Success is determined by the environment's native termination condition, i.e., whether cube 0 reaches the specified target pose.
For AntMaze-L, the script restores the Ant initial \texttt{qpos} and \texttt{qvel}, and additionally writes the first two coordinates of the future \texttt{goal\_qpos} as the internal navigation target \texttt{goal\_xy}. Success is given by the native maze termination signal, namely whether the Ant reaches the target region.
% For Puzzle, the script restores the initial \texttt{qpos}, \texttt{qvel}, and \texttt{button\_states}, uses a goal offset of 25 steps and an evaluation budget of 50 steps, and injects static goal information from the future dataset frame into the environment or wrapper. Thus, the initial button state comes from the current frame, while the target button and physical states come from the future hindsight goal; success is still determined by the environment's termination chain rather than by pixel similarity.
For Scene, the script restores the initial \texttt{qpos}, \texttt{qvel}, and \texttt{button\_states}, uses a goal offset of 50 steps and an evaluation budget of 100 steps, and binds the hindsight goal by temporarily restoring the future dataset state to extract the corresponding internal targets for cubes, buttons, drawers, and windows. The environment is then restored to the initial state before planning. Success is counted if the environment's native \texttt{terminated} signal becomes true within the evaluation budget.

\paragraph{MSE.}
Mean Squared Error (MSE) measures whether the latent representation retains physical state information that can be decoded linearly or nonlinearly.
We first freeze the trained visual encoder and map each observation $o_i$ to its latent representation $z_i=f_\theta(o_i)$;
we then train a lightweight probe $q_\psi(\cdot)$ to predict the ground-truth physical quantity $y_i$ in the environment from $z_i$,
such as cube position, end-effector position, or other task-relevant states.
The training objective of the probe is
\begin{equation}
    \psi^\star
    =
    \arg\min_{\psi}
    \frac{1}{N}
    \sum_{i=1}^{N}
    \left\|
    q_\psi(z_i) - y_i
    \right\|_2^2
\end{equation}
On the test set, MSE is computed as
\begin{equation}
    \mathrm{MSE}
    =
    \frac{1}{N d_y}
    \sum_{i=1}^{N}
    \left\|
    q_{\psi^\star}(z_i) - y_i
    \right\|_2^2
\end{equation}
where $d_y$ denotes the dimensionality of the predicted physical quantity.
A lower MSE indicates that the latent representation encodes task-relevant physical state information more faithfully.
It is worth noting that MSE does not directly measure the final planning success rate; rather, it serves as an auxiliary metric for analyzing whether the latent space learned by different methods contains decodable state information useful for control and planning.

% Describe metrics such as SR, failure rate, final goal distance, and progress violation.

\subsection{Additional Progress Correlation on Cube-S}
\label{app:cube_progress_correlation}

\begin{figure*}[!t]
\centering
\includegraphics[width=0.72\textwidth]{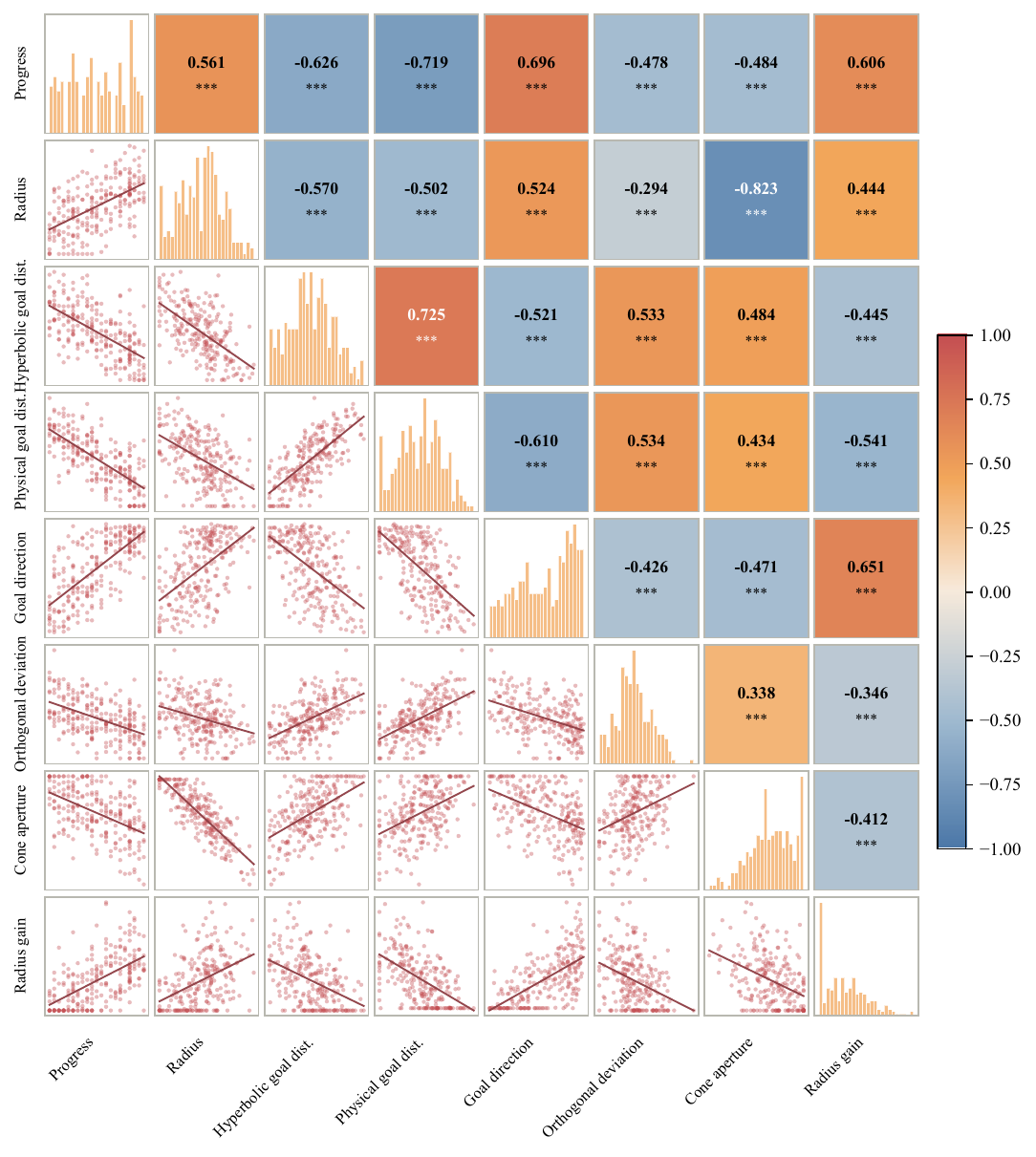}
\caption{
Spearman correlation analysis of progress-related geometric variables on Cube-S. The layout follows the PushT matrix in Fig.~\ref{fig:pusht_progress_correlation}.
}
\label{fig:cube_progress_correlation}
\end{figure*}

Fig.~\ref{fig:cube_progress_correlation} extends the progress-geometry correlation analysis to Cube-S, where the visual dynamics involve 3D robot-object interaction and cube pose changes. The overall trend is consistent with the PushT analysis: trajectory progress is positively correlated with goal-direction alignment and negatively correlated with both hyperbolic and physical goal distances, while latent radius and radius gain provide additional diagnostics of the learned coarse-to-fine organization. Compared with PushT, the correlations on Cube-S are generally weaker, which is expected because grasping, contact changes, and pose correction introduce noisier local transitions. Nevertheless, the same goal-directional organization remains visible, supporting that the learned progress geometry is not specific to the planar PushT setting.

\section{Implementation Details}
\label{app:implementation}

\subsection{Network Architecture}
\label{app:network_architecture}

The ProWorld network comprises a visual encoder, a Euclidean latent projection head, a hyperbolic projection head, an action encoder, and an action-conditioned autoregressive predictor. All pixel inputs are first normalized with ImageNet statistics and resized to $224\times224$ without offline preprocessing. The visual encoder uses a ViT-tiny backbone with scale and patch size of $14$, without pretrained weights or mask tokens. We take the CLS token from the ViT output as the visual feature and project it into the Euclidean latent space via a two-layer MLP:
\begin{equation}
    z_t = g_\psi(f_\theta(o_t)) \in \mathbb{R}^{192}
\end{equation}
The Euclidean projection head consists of Linear-BatchNorm-GELU-Linear, with a hidden dimension of $2048$ and an output dimension of $192$.

The action encoder maps environment actions to the same dimensionality as the state latent variables. For datasets with frame-skip, one model timestep corresponds to multiple underlying environment steps. If the raw action dimensionality is $d_a$ and each model timestep comprises $\texttt{frameskip}$ consecutive actions, the input dimensionality of the action encoder is $\texttt{frameskip}\times d_a$. The default is $\texttt{frameskip}=5$ for OGBench Single-Expert and $\texttt{frameskip}=1$ for AntMaze. The action encoder first applies a $1$D convolution to map the action channels to a smoothed dimension, followed by a two-layer MLP to produce the action embedding:
\begin{equation}
    \alpha_t = e_\eta(a_t) \in \mathbb{R}^{192}
\end{equation}

The dynamics predictor employs an action-conditioned Transformer. The default history window length is $H=3$ and the prediction horizon is $1$. The predictor consists of $6$ Transformer blocks with $16$ attention heads, each of dimension $64$, an MLP hidden dimension of $2048$, and dropout of $0.1$. Action conditioning is injected via an AdaLN-style conditional block. The predictor first outputs the future state prediction in the Euclidean latent space and maps it back to the $192$-dimensional latent space through a prediction projection head:
\begin{equation}
    \hat{z}_{t+1} = p_\phi(z_{t-H+1:t}, \alpha_{t-H+1:t})
\end{equation}

The hyperbolic projection head maps Euclidean latent variables to the tangent space at the origin of the Lorentz model. The default manifold dimension is $193$, comprising $192$ spatial components plus one additional Lorentz time coordinate. Accordingly, the hyperbolic projection head outputs a $192$-dimensional tangent vector:
\begin{equation}
    u_t = q_\omega(z_t) \in \mathbb{R}^{192}
\end{equation}
The corresponding hyperbolic point $h_t\in\mathbb{H}^{192}_{c}$ in the Lorentz model is then obtained via the exponential map at the origin. The hyperbolic projection head is likewise a two-layer MLP with hidden dimension $384$ and LayerNorm as the normalization layer.

\subsection{Optimization Details}
\label{app:optimization_details}

All models are trained with the AdamW optimizer. The default learning rate is $3\times10^{-5}$ and weight decay is $10^{-3}$. The learning rate schedule follows \texttt{LinearWarmupCosineAnnealingLR}, updated per epoch. Training runs for $100$ epochs in total using three random seeds, $42$, $1337$, and $3407$. We use bfloat16 precision by default and apply gradient clipping with a clip value of $1.0$. Evaluation uses the same seed as the corresponding training run.

The batch size is $128$ across all datasets. The training objective consists of four terms:
\begin{equation}
\begin{aligned}
\mathcal{L}=
&\lambda_{\mathrm{pred}}\mathcal{L}_{\mathrm{pred}}
+
\lambda_{\mathrm{sig}}\mathcal{L}_{\mathrm{sig}}\\
&+
\lambda_{\mathrm{ctr}}\mathcal{L}_{\mathrm{ctr}}
+
\lambda_{\mathrm{cone}}\mathcal{L}_{\mathrm{cone}}
\end{aligned}
\end{equation}
The default weights are
\begin{equation}
\begin{aligned}
&\lambda_{\mathrm{pred}}=1.0
\qquad
\lambda_{\mathrm{ctr}}=0.03 \\
\qquad
&\lambda_{\mathrm{cone}}=0.02
\qquad
\lambda_{\mathrm{sig}}=0.06
\end{aligned}
\end{equation}
SIGReg uses $17$ knots and $1024$ random projection directions. The contrastive learning temperature is $0.15$, and label smoothing defaults to $0$.

\subsection{Computational Overhead}
\label{app:computational_overhead}

Tab.~\ref{tab:computational_overhead} compares the parameter count and FLOPs of ProWorld against the LeWM backbone.
The additional hyperbolic components introduce only a small overhead: total parameters increase by $0.83\%$, while the rollout-5 FLOPs increase by $0.03\%$.
This indicates that the progress-aware hyperbolic geometry adds negligible computational cost compared with the visual encoder and action-conditioned dynamics predictor.

\begin{table}[t]
\centering
\small
\setlength{\tabcolsep}{3pt}
\caption{Computational overhead of ProWorld compared with LeWM.}
\label{tab:computational_overhead}
\resizebox{\linewidth}{!}{
\begin{tabular}{lrrrr}
\toprule
\textbf{Metric} & \textbf{LeWM} & \textbf{ProWorld} & \textbf{$\Delta$} & \textbf{$\Delta$ \%} \\
\midrule
Total params & 18.034M & 18.183M & 148.800K & +0.83 \\
Trainable params & 18.034M & 18.183M & 148.800K & +0.83 \\
Buffers & 8.194K & 8.195K & 1.000 & +0.01 \\
Encode FLOPs & 11.158B & 11.159B & 1.181M & +0.01 \\
Predict FLOPs & 69.339M & 69.339M & 0.000 & +0.00 \\
Rollout-5 FLOPs & 11.514B & 11.518B & 3.839M & +0.03 \\
\bottomrule
\end{tabular}
}
\end{table}

\subsection{Hyperbolic Geometry Implementation}
\label{app:hyperbolic_geometry_implementation}

ProWorld implements the hyperbolic latent space using the Lorentz model. For curvature $c>0$, the Lorentz manifold is defined as
\begin{equation}
\begin{aligned}
    \mathbb{H}^{m}_{c}
    =
    \Big\{
    x\in\mathbb{R}^{m+1}
    \quad\big|\quad
    &\langle x,x\rangle_{\mathcal{L}}=-\frac{1}{c} \\
    &x_0>0
    \Big\}
\end{aligned}
\end{equation}
where the Lorentz inner product is
\begin{equation}
    \langle x,y\rangle_{\mathcal{L}}
    =
    \sum_{i=1}^{m}x_i y_i - x_0 y_0
\end{equation}
We use a fixed curvature of $c=1.0$ by default in the main experiments. For the learnable-curvature ablation, the effective curvature is parameterized as
\begin{equation}
    c = \mathrm{softplus}(\tilde{c}) + \varepsilon_{\mathrm{num}}
\end{equation}
This ensures that the learned curvature remains strictly positive in that ablation.

Given a tangent vector $u$, we first clip its norm to a maximum of $4.0$ to prevent numerical overflow in the exponential map. The Lorentz exponential map at the origin is then applied:
\begin{equation}
    \exp^c_{\mathbf{0}}(u)
    =
    \begin{bmatrix}
    \frac{\cosh(\sqrt{c}\|u\|)}{\sqrt{c}} \\
    \frac{\sinh(\sqrt{c}\|u\|)}{\sqrt{c}\|u\|}u
    \end{bmatrix}
\end{equation}
When $\|u\|$ is very small, we clamp the denominator with $\varepsilon_{\mathrm{num}}$ to avoid division by zero. After the exponential map, a projection step re-projects the result onto the Lorentz manifold. Specifically, given the spatial components $x_{1:m}$, the time coordinate is recomputed as
\begin{equation}
    x_0 = \sqrt{\frac{1}{c} + \|x_{1:m}\|_2^2}
\end{equation}

Hyperbolic distance is computed via the Lorentz inner product:
\begin{equation}
    d_{\mathbb{H}}(x,y)
    =
    \frac{1}{\sqrt{c}}
    \operatorname{arcosh}
    \left(
    -c\langle x,y\rangle_{\mathcal{L}}
    \right)
\end{equation}
In the implementation, the input to $\operatorname{arcosh}$ is clamped from below to $1+\varepsilon_{\mathrm{num}}$ to ensure numerical validity. For bfloat16 or float16 inputs, distance computation is promoted to float32 by default to improve training stability.

The entailment cone loss is computed directly in the Lorentz model. For each parent-child-goal triple $(h^p,h^c,h_g)$, we first compute the child and goal tangent directions at the parent:
\begin{equation}
    v_c=\log^c_{h^p}(h^c),
    \qquad
    v_g=\log^c_{h^p}(h_g)
\end{equation}
The directional angle $\theta_{c,g}$ is computed between the normalized tangent directions. The cone aperture adapts to the normalized parent-to-goal distance:
\begin{equation}
 \begin{aligned}
    \delta_p^g =
    \operatorname{clip}
    \left(
    \frac{d_{\mathbb{H}}(h^p,h_g)}{\rho_{\max}},0,1
    \right),\\
    A(\delta_p^g)
    =
    A_{\min}
    +
    (A_{\max}-A_{\min})\delta_p^g
 \end{aligned}
\end{equation}
The defaults are $A_{\min}=0.10$ and $A_{\max}=0.95$. The final cone penalty includes both an angular violation term and a goal-distance progress term:
\begin{equation}
\begin{aligned}
    \mathcal{L}_{\mathrm{cone}}
    =
    \mathbb{E}
    \Big[
    &\max(0,\theta_{c,g}-A(\delta_p^g)) \\
    &+
    \max(0,d_{\mathbb{H}}(h^c,h_g)+m_g \\
    &\hspace{10mm}
    -d_{\mathbb{H}}(h^p,h_g))
    \Big]
\end{aligned}
\end{equation}
where the default goal-progress margin is $m_g=0.03$. This formulation encourages future states to move from the parent toward the hindsight goal and to reduce their hyperbolic goal distance.

\section{Hyperparameter Settings}
\label{app:hyperparameters}

Tabs.~\ref{tab:training_hparams} and~\ref{tab:planning_hparams} summarize the default training and planning configurations used in our experiments.

\subsection{Loss Weights}
\label{app:loss_weights}

The training objective of ProWorld consists of a hyperbolic prediction loss, a Lorentz contrastive learning loss, an entailment cone loss, and the SIGReg regularization.

Here, $\lambda_{\mathrm{pred}}$ controls the prediction loss in hyperbolic space, $\lambda_{\mathrm{ctr}}$ controls the Lorentz contrastive learning loss, $\lambda_{\mathrm{cone}}$ controls the entailment cone constraint, and $\lambda_{\mathrm{sig}}$ controls the SIGReg representation regularization. The temperature for contrastive learning is set to $\tau=0.15$, and label smoothing is set to $0.0$.

\subsection{Planning Parameters}
\label{app:planning_parameters}

The planning stage uses a CEM solver by default. At each planning step, CEM samples $300$ candidate action sequences and performs $30$ optimization iterations. In each round, CEM retains the top-$k$ candidates, where $k=30$. The initial variance scaling factor is $\mathrm{var\_scale}=1.0$.

The hyperbolic planning cost consists of the terminal distance, the best intermediate distance, and the mean trajectory distance:
\[
C =
\beta_T C_T
+
\beta_b C_{\mathrm{best}}
+
\beta_m C_{\mathrm{mean}}
\]
where $C_T$ denotes the hyperbolic distance from the predicted terminal state to the goal, $C_{\mathrm{best}}$ denotes the cost of the intermediate state along the trajectory closest to the goal, and $C_{\mathrm{mean}}$ denotes the mean goal distance over the entire predicted trajectory.

\subsection{Entailment Cone Parameters}
\label{app:cone_parameters}

The entailment cone constraint models the hierarchical progress relationship between earlier and later states with respect to a hindsight goal. Given a parent state $h_t$, a child state $h_{t+\Delta}$, and a goal state $h_g$ from the same goal-conditioned hindsight segment, the default temporal offset is $\Delta=2$. In the implementation, $\Delta$ is clipped to a valid range, i.e., at least $1$ and no greater than the current sequence length minus one.

A parent point farther from the goal has a wider cone aperture, while one closer to the goal has a narrower aperture. The angular term encourages child states to fall within the parent-to-goal cone; the goal-distance term encourages child states to become closer to the hindsight goal than their parents. In this way, the model learns not only state similarity but also a goal-dependent reachability structure that progressively unfolds along the temporal direction in hyperbolic space.

\section{Qualitative Results}
\label{app:qualitative_results}
\begin{figure*}[t]
    \centering
    \includegraphics[height=0.88\textheight]{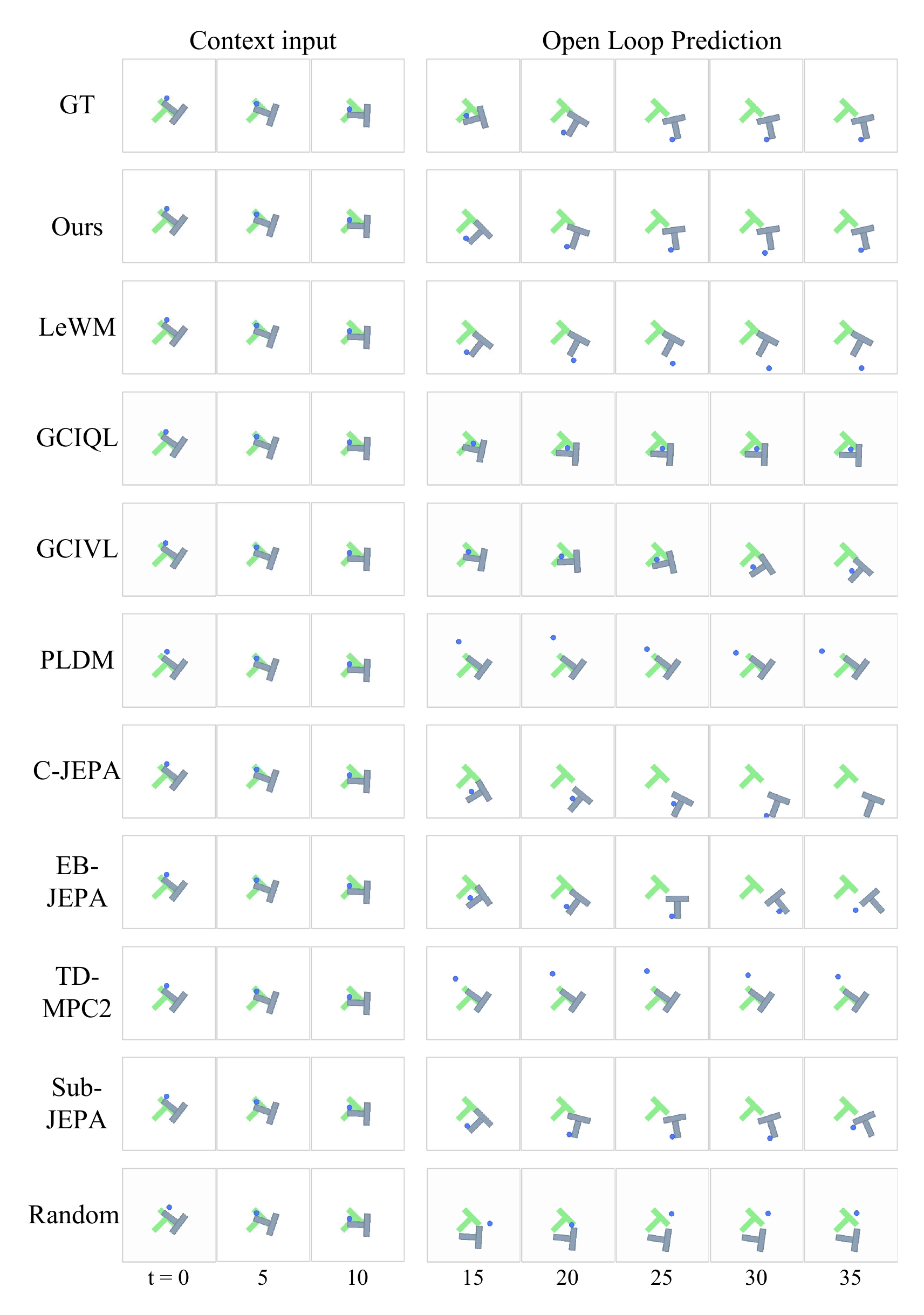}
    \caption{Open-loop prediction visualizations on PushT. The left column shows the context input, i.e., the ground-truth historical frames visible to the model before forecasting; the right column shows the open-loop prediction, i.e., future states obtained by recursively unrolling the model from the same context under a given action sequence. The top and bottom rows correspond to ground-truth observations and model-imagined results, respectively, enabling a comparison of whether ProWorld's short-term object motion predictions preserve spatial trends consistent with the true trajectory.}
    \label{fig:appendix_pusht}
\end{figure*}

Fig.~\ref{fig:appendix_pusht} presents open-loop prediction visualizations on PushT. The figure provides a side-by-side comparison between ground-truth context frames and the model's imagined rollout, serving as a qualitative check of whether the learned latent dynamics can sustain short-term object motion trends without environment feedback.

\begin{figure*}[t]
    \centering
    \includegraphics[height=0.88\textheight]{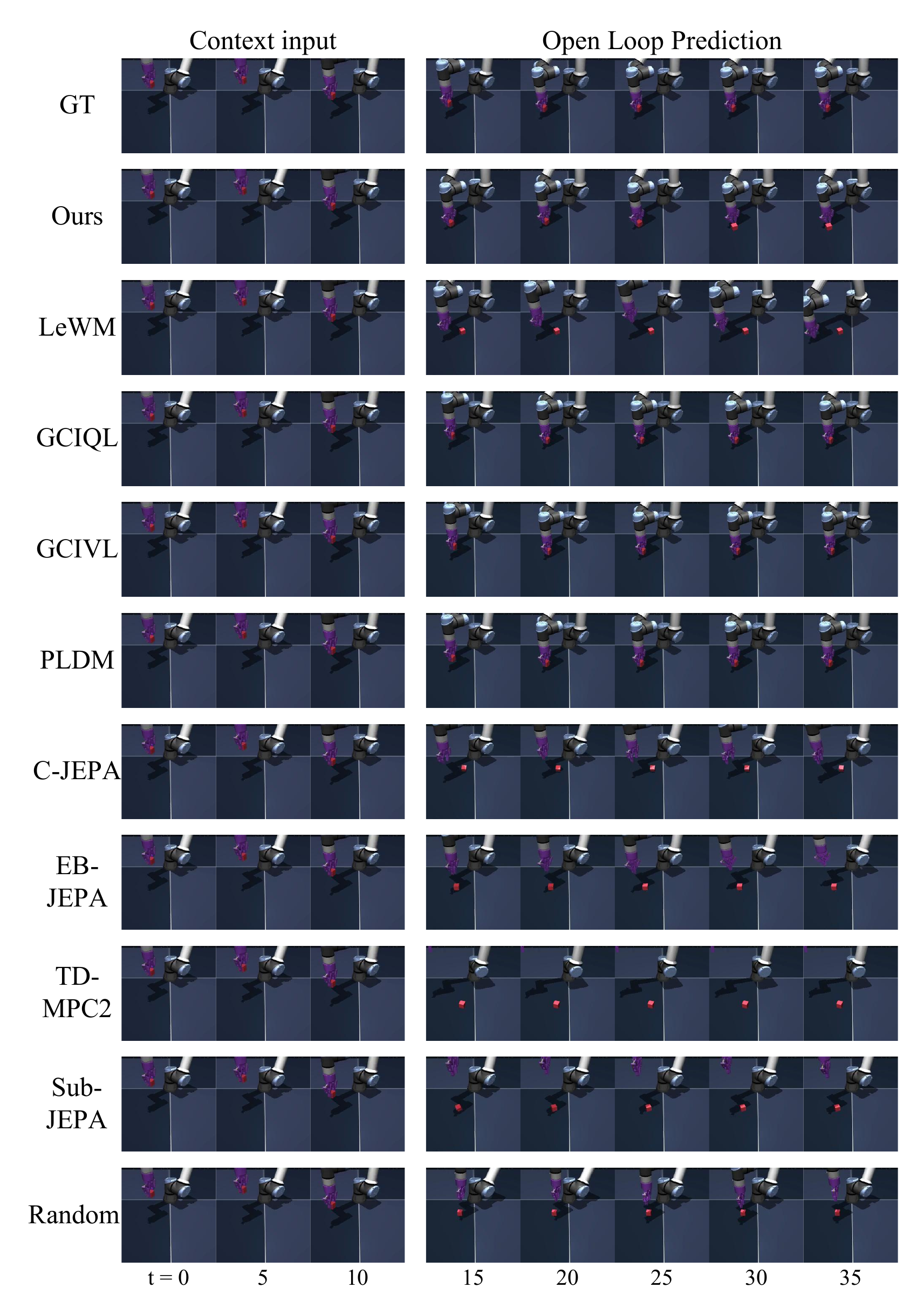}
    \caption{Open-loop prediction visualizations on Cube-S. The left block shows the context frames, and the right block shows future frames obtained by recursively unrolling each model under the same action sequence. Rows compare the ground-truth trajectory with ProWorld and baseline methods. In the Cube-S evaluation, success is recorded once the environment termination signal is triggered at any step within the evaluation budget, rather than only at the final rendered frame. Therefore, in some visualized rollouts, the cube may already have reached the target pose and then be released or dropped in later frames as the fixed-budget rollout continues. Such post-success drift reflects the visualization protocol after the success condition has been met and does not invalidate the earlier successful achievement of the goal.}
    \label{fig:appendix_cube}
\end{figure*}

Fig.~\ref{fig:appendix_cube} further extends this qualitative comparison to Cube-S, where the visual dynamics involve 3D robot-object interaction rather than planar object pushing. Compared with the baselines, ProWorld preserves the grasp-and-transport trend more closely over the open-loop horizon, while several alternatives show earlier deviations in cube pose or contact configuration. This visual evidence complements the quantitative SR results by illustrating how progress-aware latent dynamics can better maintain task-relevant manipulation structure during recursive prediction.

% \subsection{Rollout Visualization}
% \label{app:rollout_visualization}

% \textcolor{red}{Visualize latent rollout or image-space trajectories for different methods.}

% \subsection{Radial Progress in Hyperbolic Space}
% \label{app:radial_progress}

% \textcolor{red}{Analyze whether the radial displacement of states in hyperbolic space conforms to the progress hierarchy.}

% \subsection{Goal-Distance Monotonicity}
% \label{app:goal_distance_monotonicity}

% \textcolor{red}{Analyze whether the predicted trajectory's distance to the goal decreases over time.}

% \subsection{Cone Consistency}
% \label{app:cone_consistency}

% \textcolor{red}{Analyze whether goal states or successor states fall within the progress cone induced by the current state.}

\section{Training Curves}
\begin{figure*}[t]
    \centering
    \begin{subfigure}[t]{0.74\textwidth}
        \centering
        \includegraphics[width=\linewidth]{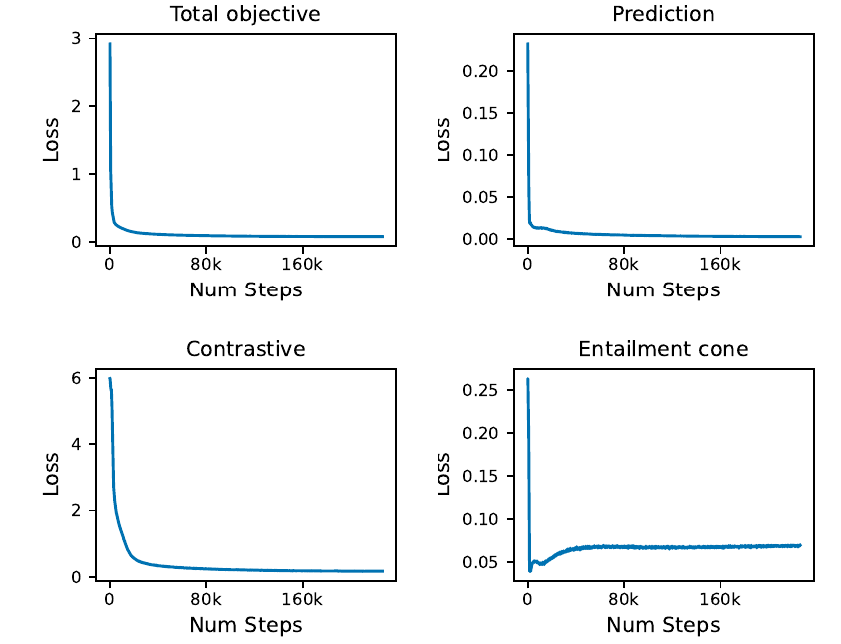}
        \caption{Representative training losses.}
        \label{fig:training_losses}
    \end{subfigure}
    \vspace{1mm}
    \begin{subfigure}[t]{0.96\textwidth}
        \centering
        \includegraphics[width=\linewidth]{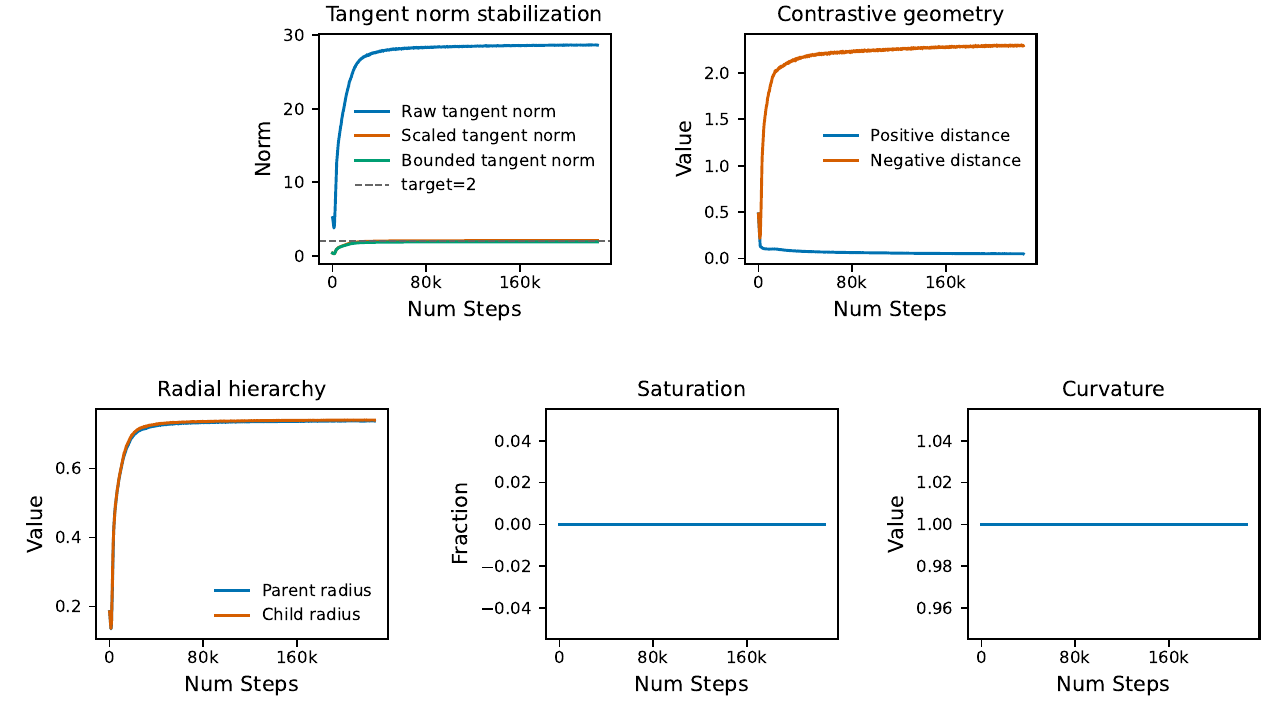}
        \caption{Hyperbolic geometry diagnostics.}
        \label{fig:training_diagnostics}
    \end{subfigure}
    \caption{Training behavior of ProWorld. (a) shows the total objective, hyperbolic prediction, Lorentz contrastive learning, and entailment cone losses. (b) shows tangent-space norm stabilization, hyperbolic positive-negative distance separation, parent-child radial organization, tangent vector saturation ratio, and the curvature parameter. These curves indicate that the prediction-related and progress-structure terms stabilize during training while the hyperbolic latent geometry remains numerically well behaved.}
\end{figure*}

Fig.~\ref{fig:training_losses} and Fig.~\ref{fig:training_diagnostics} report representative training losses and geometric diagnostic metrics, respectively. The former tracks the total objective and the main prediction, contrastive, and entailment terms used in ProWorld; the latter further tracks hyperbolic geometric quantities including tangent-space norm stabilization, contrastive distance separation, radial organization, saturation, and curvature, confirming that the hyperbolic latent space does not undergo numerical collapse during training.

\section{Limitations and Future Work}
\label{sec:limitations}
Several directions remain worthy of further investigation. First, although the cone constraint is anchored by the hindsight goal, progress pairs are still constructed primarily from temporal ordering within goal-conditioned hindsight trajectory segments; for tasks involving backtracking, detours, or complex sub-goal switching, the progress supervision may still be relatively coarse-grained. Second, the hyperbolic latent space and progress constraints introduce additional geometric modeling choices, and their stability in more complex and open-ended visual interaction settings remains to be further verified. Future work could explore more adaptive progress estimation methods and further integrate progress-aware world models with online exploration, hierarchical planning, or policy learning.

\end{document}